\documentclass[11pt, logo, onecolumn, copyright, colorlinks=true, allcolors=blue]{nvidiatechreport}
\usepackage{graphicx}
\usepackage{subcaption,ragged2e}
\usepackage[round]{natbib}
\usepackage[normalem]{ulem}
\usepackage[utf8]{inputenc}
\usepackage[T1]{fontenc}
\usepackage{url}
\usepackage{tikz}
\usepackage{enumitem}
\usetikzlibrary{positioning}
\usetikzlibrary{calc}
\usepackage{array}
\usepackage{booktabs}
\usepackage{wrapfig}
\usepackage{caption}
\usepackage{hyperref}
\usepackage{listings}
\usepackage{adjustbox}
\usepackage{footnote}
\usepackage{multirow}
\usepackage{amsmath}
\usepackage{amsfonts}
\usepackage{breakcites}
\usepackage{blindtext}
\usepackage{svg}
\usepackage{setspace}
\usepackage[most]{tcolorbox}
\usepackage{tabularx}
\usepackage{graphicx} 
\title{Pretraining Large Language Models with NVFP4}
\author{\large NVIDIA}
\date{}

\begin{document}

\begin{abstract}

\textbf{Abstract.} Large Language Models (LLMs) today are powerful problem solvers across many domains, and they continue to get stronger as they scale in model size, training set size, and training set quality, as shown by extensive research and experimentation across the industry. Training a frontier model today requires on the order of tens to hundreds of yottaflops, which is a massive investment of time, compute, and energy. Improving pretraining efficiency is therefore essential to enable the next generation of even more capable LLMs.
While 8-bit floating point (FP8) training is now widely adopted, transitioning to even narrower precision, such as 4-bit floating point (FP4), could unlock additional improvements in computational speed and resource utilization. However, quantization at this level poses challenges to training stability, convergence, and implementation, notably for large-scale models trained on long token horizons.

\smallskip
In this study, we introduce a novel approach for stable and accurate training of large language models (LLMs) using the NVFP4 format. Our method integrates Random Hadamard transforms (RHT) to bound block-level outliers, employs a two-dimensional quantization scheme for consistent representations across both the forward and backward passes, utilizes stochastic rounding for unbiased gradient estimation, and incorporates selective high-precision layers. We validate our approach by training a 12-billion-parameter model on 10 trillion tokens -- the longest publicly documented training run in 4-bit precision to date. Our results show that the model trained with our NVFP4-based pretraining technique achieves training loss and downstream task accuracies comparable to an FP8 baseline. For instance, the model attains an MMLU-pro accuracy of 62.58\%, nearly matching the 62.62\% accuracy achieved through FP8 pretraining.  
These findings highlight that NVFP4, when combined with our training approach, 
represents a major step forward in narrow-precision LLM training algorithms. 

\smallskip
\textbf{Code}: \href{https://github.com/NVIDIA/TransformerEngine/pull/2177}{Transformer Engine support for NVFP4 training}.
\end{abstract}

\maketitle

%%%%%%%%%%%%%%%%%%%%%%%%%%%%%%%%%
\section{Introduction}

The rapid expansion of large language models (LLMs) has increased the demand for more efficient numerical formats to lower computational cost, memory demand, and energy consumption during training. 8-bit floating point (FP8 and MXFP8) has emerged as a popular data type for accelerated training of LLMs~\citep{fp8,liu2024deepseekv3,mxfp8recipes}. Recent advances in narrow-precision hardware~\citep{nvidia2024blackwell} have positioned 4-bit floating point (FP4) as the next logical step~\citep{trainingllmswithmxfp4,fp4alltheway,optimizinglargelanguagemodeltrainingusingfp4,oscillationreducedmxfp4,quartet,efficientpretrainingexploringfp4,microscaling}, delivering a two- to three-fold boost in arithmetic performance and reducing memory usage by half compared to FP8.

This technical report presents an in-depth analysis of large language model (LLM) pretraining using NVFP4~\citep{nvfp4inference}, a 4-bit data format that extends the ``microscaling'' approach~\citep{microscaling}. Unlike 4-bit microscaling formats such as MXFP4~\citep{microscaling,ocp}, NVFP4 employs a smaller micro-block structure, which more effectively captures the local dynamic range in the data. NVFP4 also utilizes an FP8 scale factor format that incorporates fractional precision for more accurate microscaling. 
In addition, NVFP4 employs a two-level scaling strategy, which combines a fine-grained FP8 scale factor with an FP32 scale applied at the tensor level. These design choices allow for more precise and accurate representation of tensor values during training.

Leveraging the NVFP4 format, we introduce a 4-bit training methodology that achieves accuracies comparable to FP8 on very strong language models. This approach preserves numerically sensitive layers in higher precision, utilizes two-dimensional (2D) block scaling to maintain same quantized representations across forward and backward passes, applies Random Hadamard transforms~\citep{trainingllmswithmxfp4,quartet} to disperse large-magnitude outliers, and employs stochastic rounding~\citep{trainingllmswithmxfp4,fp4alltheway,oscillationreducedmxfp4,quartet,dorefa_2016} on gradients to reduce quantization bias. Ablation studies confirm that each component of this methodology is important for 4-bit training, especially in large-scale models and during long token horizons. 

To validate our approach, we train a very strong 12-billion parameter LLM~\citep{nemotronnano2} on 10 trillion tokens, demonstrating that its loss curve and accuracies on downstream tasks closely match with those of an FP8 baseline. While our work establishes the feasibility of FP4 training at large scales, this report is primarily concerned with the underlying algorithms and methodology rather than with runtime efficiency or system-level optimizations. This marks, to our knowledge, the first successful demonstration of training billion-parameter language models with 4-bit precision over a multi-trillion-token horizon, laying the foundation for faster and more efficient training of future frontier models.

The remainder of this technical report is organized as follows: Section~\ref{sec:nvfp4} describes the NVFP4 format, Section~\ref{sec:12btrain} presents results for a 12 billion model trained on 10 trillion tokens with NVFP4, Section~\ref{sec:recipe} discusses the training methodology for NVFP4, and Section~\ref{sec:mxfp4} compares training with NVFP4 and MXFP4. The appendices include details of the training setup (models, datasets, and hyperparameters), the quantization procedure, and ablation studies analyzing the impact of different technique choices.

%%%%%%%%%%%%%%%%%%%%%%%%%%%%%%%%%
\section{NVFP4 Format}\label{sec:nvfp4}

Due to the limited range of narrow floating-point formats, microscaling (MX) formats~\citep{ocp} were introduced to balance dynamic range and precision. These formats are characterized by a block-wise representation where a group of data elements shares a single, common scale factor. MX formats include 8-bit (MXFP8), 6-bit (MXFP6), and 4-bit (MXFP4) floating-point types. In MXFP4, each element is represented as E2M1\footnote{Floating-point types are denoted as E$x$M$y$ and consist of one sign bit, $x$ exponent bits, and $y$ mantissa bits.}~\citep{ocp}, meaning it has 1 sign bit, 2 exponent bits, and 1 mantissa bit. This allows MXFP4 to encode the values $\pm0$, $\pm0.5$, $\pm1$, $\pm1.5$, $\pm2$, $\pm3$, $\pm4$, and $\pm6$.

Since original higher-precision values (e.g., FP32 or BF16) often exceed the FP4 range, they must be scaled into the representable range during quantization. Scale factors are typically chosen so that the absolute maximum value (amax) within a block maps to the FP4 maximum representable, favoring the prevention of saturations while minimizing small magnitudes being lost to zero. After scaling, high precision values in a tensor are rounded to the nearest FP4-representable number and later decoded back to their original range 
using the reciprocal of the same scale. To 
improve hardware efficiency, MX formats store block scale factors in 8 bits. Each block of 32 contiguous elements in a tensor shares a single 8-bit scale factor, stored in an unsigned E8M0 format (UE8M0), which encodes a power-of-two value ranging from $2^{-127}$ to $2^{127}$.~\cite{mxfp8recipes} found that it is beneficial to round scale factors up to the next representable UE8M0 value to avoid saturations.

\begin{figure}[ht] \centering
    \includegraphics[width=0.65\textwidth]
    {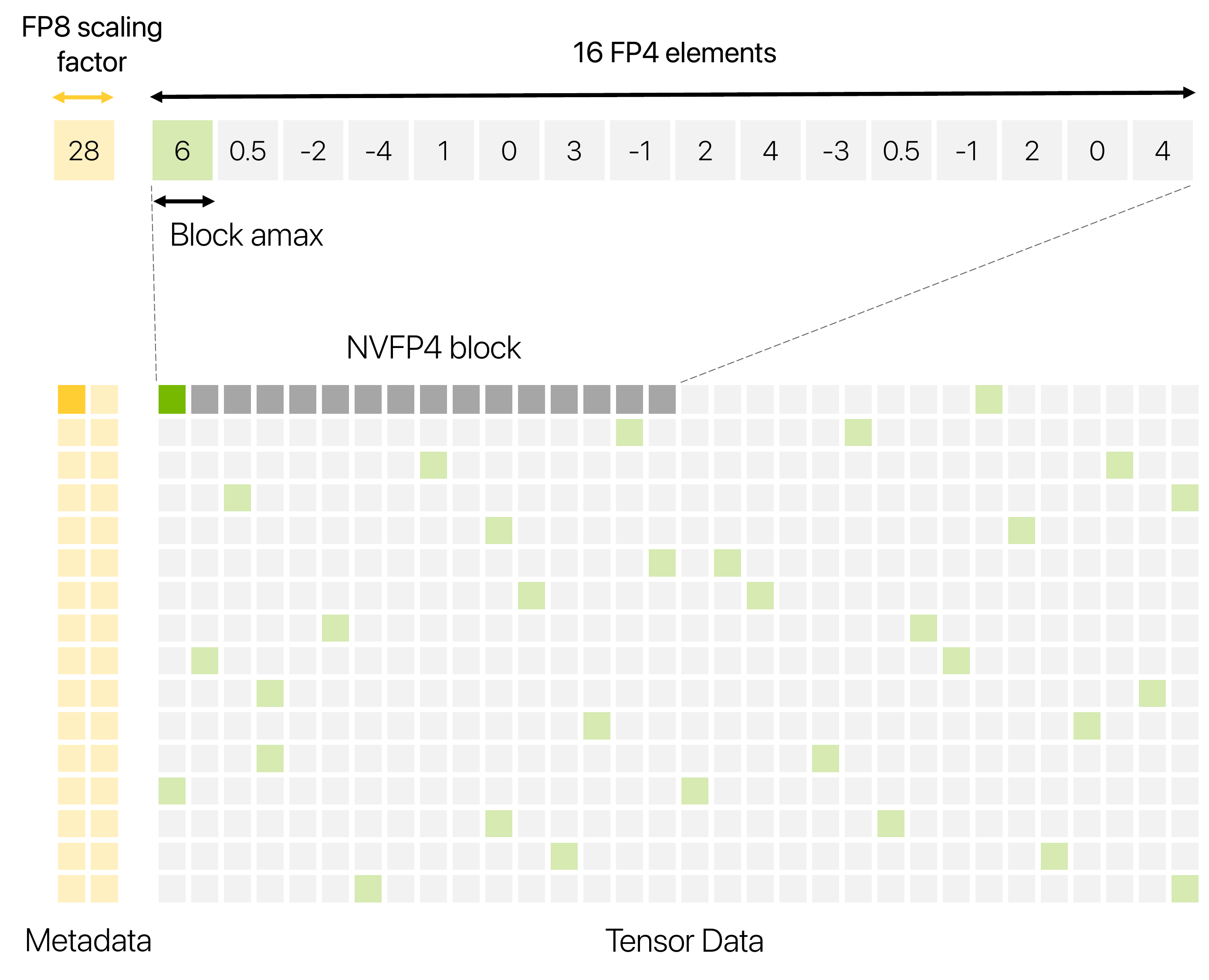}
    \caption{A 16$\times$32 matrix stored in NVFP4 format. Each block contains sixteen contiguous FP4 elements (gray and green) along with a single FP8 scale factor (yellow). The element with the largest magnitude in each block (green) is scaled to the FP4 maximum representable value and can be recovered using the block scale factor. A per-tensor FP32 scale factor (not shown) is also applied.} 
    \label{fig:nvfp4}
 \end{figure}

NVFP4 is an enhanced 4-bit format that provides improved numerical properties over MXFP4. First, by reducing  the block size from 32 to 16 elements, NVFP4 narrows the dynamic range within each block, better fitting values into the FP4 range. Second, block scale factors are stored in E4M3 rather than UE8M0, trading some exponent range for additional mantissa bits. Third, an FP32 scale is applied at the tensor level to retain the range of block scales. With such a two-level microscaling approach, NVFP4 encodes at least 6.25\% of values in a block (the amax values  in each block of $16$ elements) at near-FP8 precision, while storing the remaining values in FP4 (see Figure~\ref{fig:nvfp4}). In contrast, MXFP4 stores all values in FP4, and can potentially lose up to one binade of dynamic range (and four samples: $\pm4$ and $\pm6$) because of power-of-two scale factor rounding (see Appendix~\ref{app:mxfp4scale} for details).

For NVFP4, having more precise scaling with E4M3 reduces the range available for representing the scale factors. 
As a result, a second level of FP32 scaling is used to adjust the original tensor's distribution such that block scale factors can be represented in E4M3. This two-level scaling scheme works as follows: (1) a per-tensor FP32 scale remaps all the values within a tensor into representable range of a block (FP4 $\times$ FP8), then (2) a per-block E4M3 scale moves the values within a block into FP4 representable range. Appendix~\ref{app:quantization} describes the quantization and scaling strategy in more detail.

In summary, NVFP4’s design improvements over MXFP4 increase the accuracy of outliers while minimizing the amount of small values being quantized to zero. These numerical advances (smaller block size and more precise scaling) give NVFP4 a clear advantage over MXFP4, resulting in consistently better training behavior. We discuss training results comparing these two formats in Section~\ref{sec:mxfp4}.

\begin{table}[ht]
\centering
\caption{NVIDIA Blackwell Tensor Cores.}\label{tab:tensorcore}
\begin{tabular}{lcccccc}
\toprule
\textbf{Format} & \textbf{Element} & \textbf{Scale} & \textbf{Block} & \multicolumn{2}{c}{\textbf{Speedup vs. BF16}} \\
\cmidrule(lr){5-6}
& & & & \textbf{GB200} & \textbf{GB300} \\
\midrule

MXFP8  & E5M2/E4M3 & UE8M0 & 32    & 2$\times$ & 2$\times$ \\
MXFP6  & E3M2/E2M3 & UE8M0 & 32    & 2$\times$ & 2$\times$ \\
MXFP4  & E2M1      & UE8M0 & 32    & 4$\times$ & 6$\times$ \\
NVFP4  & E2M1      & E4M3 &  16 & 4$\times$ & 6$\times$ \\

\bottomrule
\end{tabular}
\end{table}

\textbf{Hardware support via Tensor Cores:}\label{sec:tensorcores}
NVIDIA Blackwell GPUs provide native support for general matrix multiplications (GEMMs) for a wide range of microscaling formats -- MXFP8, MXFP6, MXFP4, NVFP4 -- as summarized in Table~\ref{tab:tensorcore}. Tensor Cores read narrow precision inputs along with 8-bit scale factors for each block of \(16\) or \(32\) elements. Tensor Cores compute partial dot-products over the block, multiply each partial product by the corresponding scale factors to descale the inputs scaled during quantization, and accumulate the partial results in higher precision to produce the final dot-product in FP32. 
Further, Blackwell GPUs have native support for several rounding modes including round-to-nearest-even and stochastic rounding for FP4 conversion instructions.
Tensor Cores deliver FP4 computations at 2$\times$ (on GB200 chips) and 3$\times$ (on GB300 chips) higher math throughput rates compared to FP8. Memory usage is also approximately halved when using FP4 operands compared to FP8.
As a result, FP4 could offer significant speedups for LLM training when GEMMs make up a substantial portion of training time.

%%%%%%%%%%%%%%%%%%%%%%%%%%%%%%%%%
\section{Training with NVFP4}\label{sec:12btrain}

We report training results for a 12B-parameter hybrid Mamba-Transformer model trained on 10T tokens with NVFP4 precision and compare the results against an FP8 reference model.

\textbf{Model and training setup:}\label{sec:setup}
We consider a hybrid Mamba-Transformer model architecture used in the recently introduced Nemotron-H family of models~\citep{nemotronnano2,nemotronh2025}. These models consist of a mixture of Mamba-2, Self-Attention, and FFN blocks. 
We use the same architecture as the Nemotron-Nano-12B-v2-Base model (a 12B-parameter model from the Nemotron-H family~\citep{nemotronnano2}), which has been shown to achieve competitive accuracies across multiple benchmarks. We train this model on 10T tokens with a Warmup-Stable-Decay~\citep{hu2024minicpm} learning rate  schedule, where the learning rate is constant through the first 80\% of training and then decayed over the last 20\%. Appendix~\ref{12bmodelarch} has more details on the model configuration.

We pretrain the model in NVFP4 using the methodology described in Section~\ref{sec:recipe}. To compare the loss and accuracies on downstream tasks, we pretrain an FP8 baseline following the methodology in~\citep{liu2024deepseekv3,nemotronnano2}.

\begin{figure}[ht] \centering
    \includegraphics[width=0.675\textwidth]
    {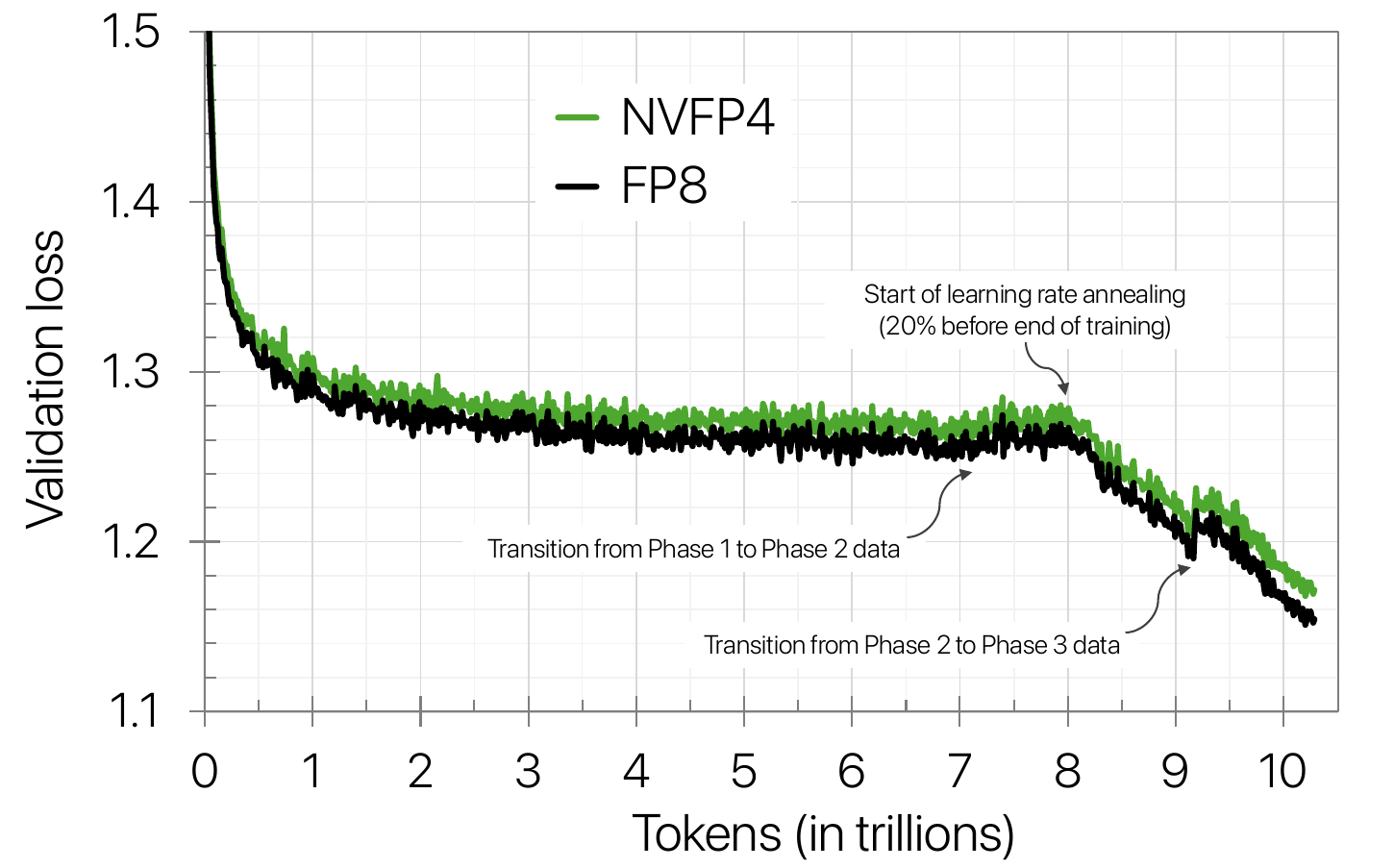}
    \caption{Validation loss of NVFP4 and FP8 pretraining for the 12B model using 10T tokens.} 
    \label{fig:loss}
\end{figure}

\begin{figure}[ht] \centering
    \includegraphics[width=0.75\textwidth]
    {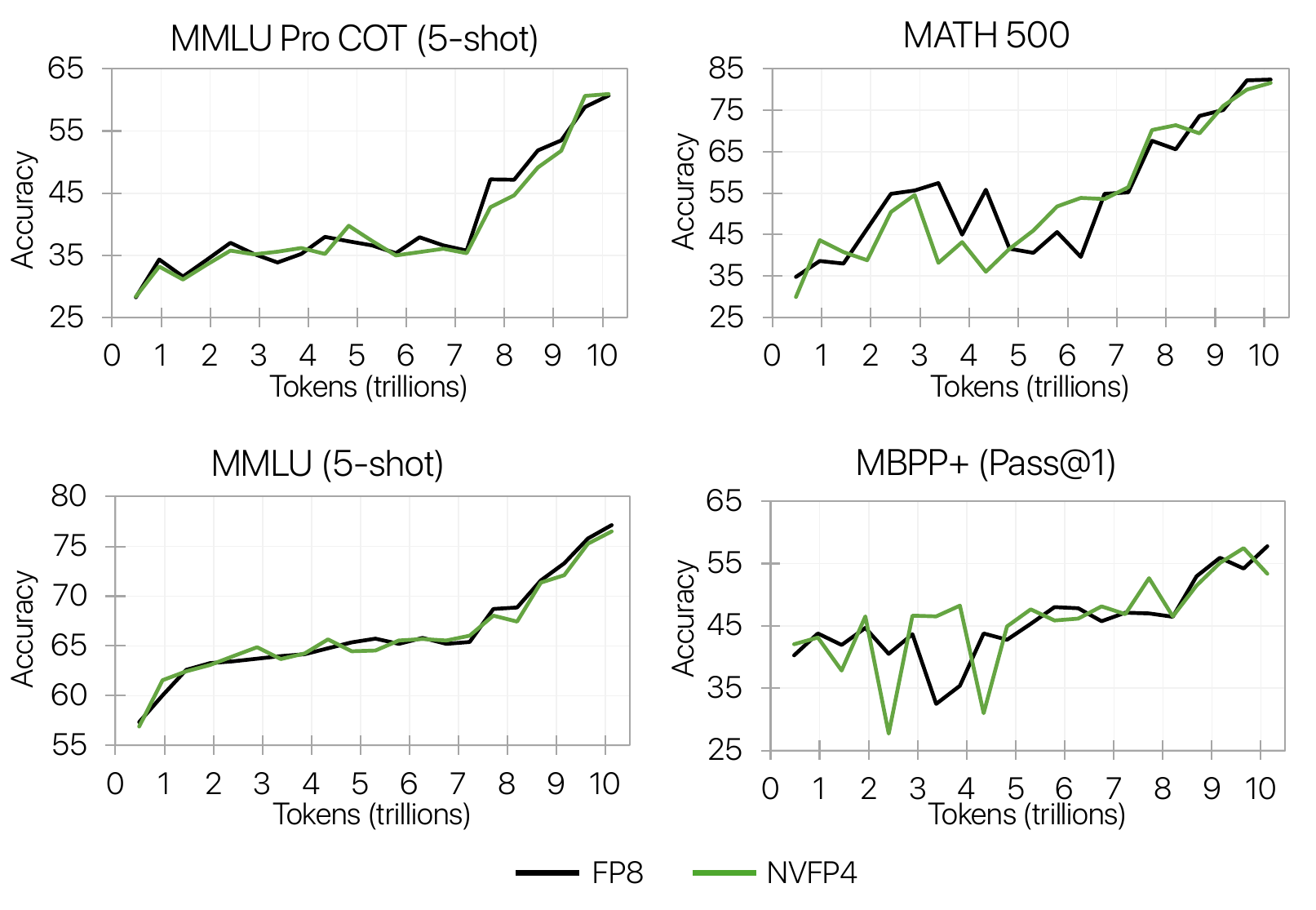}
    \caption{Task accuracy of NVFP4 versus FP8 measured throughout 10T tokens of pretraining.} 
    \label{fig:task_scores}
\end{figure}

\textbf{Pretraining results:}
\label{sec:results}
Figure~\ref{fig:loss} shows that the validation loss of NVFP4 closely tracks its FP8 counterpart throughout training. During the stable phase of training, the relative loss error of NVFP4 remains consistently below $1\%$, and widens to slightly above $1.5\%$ as the learning rate is decayed towards the end of training. This indicates that the training dynamics of NVFP4 closely follows FP8, with only a small divergence appearing late in training. Note that the change in the slope of the loss curve at 8T tokens stems from the learning rate decay. Additionally, the small jump in loss at 9T tokens corresponds to the change in the dataset blend. Appendix~\ref{12bmodelarch} has more details on the used dataset blend.

Despite the small gap in loss, downstream task accuracies remain largely unaffected. Figure~\ref{fig:task_scores} shows NVFP4 matching FP8 on downstream evaluations over the duration of training. This trend holds across a wide range of domains, including knowledge-intensive reasoning, mathematics, coding, and commonsense reasoning tasks. Table~\ref{tab:12bevals} provides a more comprehensive view, confirming that NVFP4 achieves comparable accuracy to FP8 across most individual benchmarks. The exception is the coding task, where NVFP4 falls slightly behind. We suspect the difference could be due to noisy evaluations:  MBPP+ accuracy drops on the very final checkpoint evaluation and choosing another checkpoint could potentially lead to better accuracy for this task.

In scenarios where minimizing loss is critical, the gap can be reduced by transitioning to higher precision during the final stages of training. In particular, changing precision from NVFP4 to BF16 (or potentially, MXFP8) during the decay phase mitigates the loss gap, as explained later in Appendix~\ref{app:healing}. This implies most of the training can be executed in NVFP4 (with a small amount of training in higher precision) to achieve losses that are closer to the FP8 baseline.

These results confirm that NVFP4 training remains stable over long token horizons, preserving accuracy relative to higher-precision baselines, and demonstrate that our NVFP4 training methodology offers a practical pathway for scalable 4-bit training.

\begin{table*}[ht]
\centering
\caption{Accuracy of the 12B model for FP8 and NVFP4 pretraining. Evaluations are done in BF16.}
\label{tab:12bevals}
\small
\begin{minipage}{0.48\linewidth}
\centering
\begin{tabular}{lcc}
\toprule
\textbf{Task} & \textbf{FP8} & \textbf{NVFP4} \\
\midrule
\textbf{General}      & \textbf{68.99} & \textbf{69.82} \\
MMLU                  & 77.36 & 76.57 \\
MMLU-Pro 5-shot       & 62.62 & 62.58 \\
AGIEval English CoT   & 67.01 & 70.31 \\
\midrule
\textbf{Math}         & \textbf{86.20} & \textbf{86.88} \\
GSM8k CoT             & 89.08 & 92.27 \\
MATH                  & 83.32 & 81.48 \\
\midrule
\textbf{Multilingual} & \textbf{77.93} & \textbf{80.24} \\
Global MMLU           & 74.00 & 74.94 \\
MGSM                  & 81.87 & 85.53 \\
\bottomrule
\end{tabular}
\end{minipage}
\hfill
\begin{minipage}{0.48\linewidth}
\centering
\begin{tabular}{lcc}
\toprule
\textbf{Task} & \textbf{FP8} & \textbf{NVFP4} \\
\midrule
\textbf{Code}       & \textbf{59.52} & \textbf{56.67} \\
HumanEval$+$        & 59.93 & 57.43 \\
MBPP$+$             & 59.11 & 55.91 \\
\midrule
\textbf{Commonsense Understanding} & \textbf{77.29} & \textbf{76.75} \\
ARC Challenge      & 91.81 & 91.81 \\
HellaSwag          & 83.83 & 83.09 \\
OpenBookQA         & 47.60 & 47.40 \\
PIQA               & 82.64 & 82.70 \\
Winogrande         & 80.58 & 78.77 \\
\bottomrule
\end{tabular}
\end{minipage}
\end{table*}

\begin{figure}[thbp]
    \centering
    \includegraphics[
         width=0.70\textwidth,
    ]{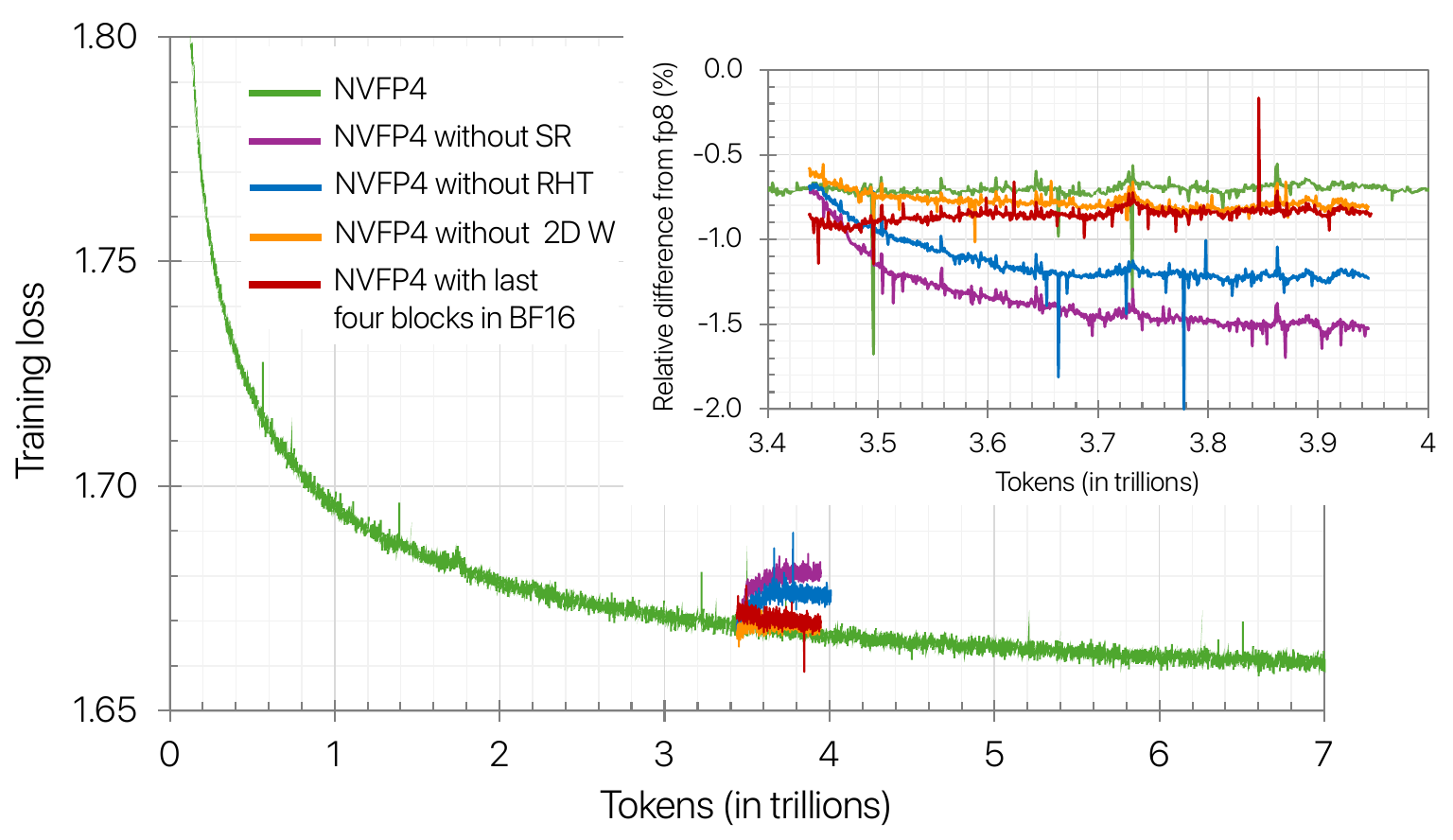}
    \caption{Ablations on the 12B model trained for 10T tokens. Ablation studies start from the model trained up to 3.43T tokens using NVFP4 except in the first two and last eight blocks, and systematically remove one methodology component at a time: stochastic rounding (SR), Random Hadamard Transforms (RHT), two-dimensional scaling (2D), and fewer blocks in BF16. Relative difference is defined as (FP8 - experiment) $/$ FP8, where a negative difference means the experiment is worse. 
    } 
    \label{fig:ablations12b}
\end{figure}

%%%%%%%%%%%%%%%%%%%%%%%%%%%%%%%%%
\section{Training Methodology}\label{sec:recipe}

In addition to the NVFP4 data type, our approach incorporates several key techniques to enable effective 4-bit training. These include (1) retention of specific numerically sensitive layers in higher precision, (2) Random Hadamard transforms to manage block-level outliers, (3) two-dimensional (2D) block scaling applied to weights for consistency between forward and backward passes, and (4) stochastic rounding to ensure unbiased quantized gradients. While smaller models trained on shorter token horizons may not require all of these techniques, we find that each component is essential for ensuring convergence and stability of the 12B model training over the 10T-token horizon. Figure~\ref{fig:ablations12b} illustrates this via ablation studies: starting with the full training methodology described below, we remove one component at a time and observe that eliminating any of them leads to worse convergence. 

In short, our recommendation for NVFP4 training is:

\begin{enumerate}
    
    \item Keep a few sensitive linear layers in higher precision (15\% of the network, with the majority of high precision layers at the end of the network).
    
    \item Apply Random Hadamard transforms of size 16$\times$16 to inputs of weight gradient GEMMs.
    
    \item Use two-dimensional (2D) scaling over 16$\times$16 blocks for weights, and one-dimensional scaling over 1$\times$16 blocks for activations and gradients.
    
    \item Use stochastic rounding for gradients and round-to-nearest-even for weights and activations.
    
\end{enumerate}

In the rest of this section, we discuss each component of the training methodology in detail and describe the ablation study presented in Figure~\ref{fig:ablations12b}. Additional ablations %at smaller scales 
are reported in Appendix~\ref{sec:appendix_ablations} to support our choices.

\subsection{Mixed precision}\label{sec:bf16layers}
\begin{figure}[htbp] \centering
    \includegraphics[
         width=0.695\textwidth,
    ]{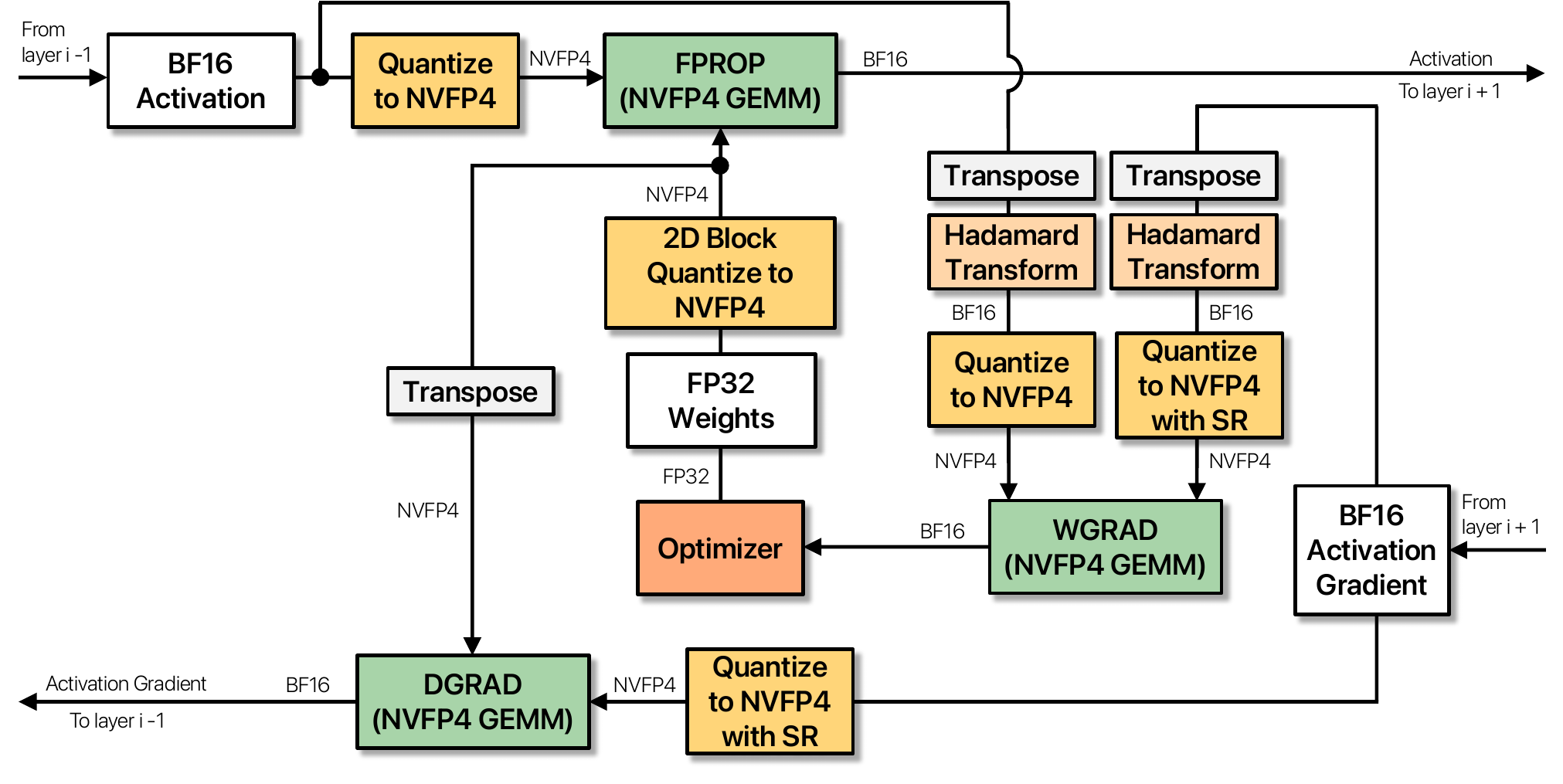}
    \caption{Illustration of compute flow for a NVFP4 quantized linear layer. All GEMM operations quantize their inputs to NVFP4.}
    \label{fig:gemm}
\end{figure}

We adopt a mixed-precision strategy for FP4 training. The majority of computations, specifically the GEMM operations within linear (fully-connected) layers, are carried out in FP4. As illustrated in Figure~\ref{fig:gemm}, each linear layer has three underlying GEMMs: a GEMM in the forward pass (Fprop), and separate GEMMs to compute activation gradients (Dgrad) and weight gradients (Wgrad) in the backward pass. GEMM operations consume FP4 tensors as inputs and produce outputs in BF16 or FP32.

\textbf{Linear layers:}
Although linear layers are typically computed in narrower precisions, we observe that some linear layers are more sensitive to FP4 than others. In particular, training diverges when every linear layer is quantized to FP4. We observe from our ablation studies (see Appendix~\ref{app:layersensitivity}) that the final few linear layers in our models cause training to diverge, since they require more dynamic range and mantissa than FP4 provides. Based on these findings, we recommend leaving a small fraction of the final layers (e.g., fewer than $15\%$) in BF16 or MXFP8 for better training convergence.

For the 12B model, we chose a conservative configuration, keeping the first two blocks in addition to the final eight blocks (FFNs or Mamba-2, each has 2 linear layers) in BF16, representing $16\%$ of the linear layers in the network in high precision. However, Figure~\ref{fig:ablations12b} indicates that convergence remains stable even when only the final four blocks are left in higher precision, suggesting that a larger portion of the model could have been safely trained in FP4.

\textbf{Attention, embedding, non-linear layers, and other tensors:}
To ensure numerical stability during training, we retain 
the original precision (e.g., BF16 or FP32) for embeddings, the output projection head, normalization layers, non-linearities, and attention components, including softmax and the query-key and attention score-value batched GEMMs. The main weights (stored by the optimizer), weight gradients (used for gradient accumulation across microbatches and across data-parallel replicas), and optimizer states are also kept in FP32. Tensor parallel reductions are performed in BF16 precision.

\subsection{Random Hadamard Transforms}\label{sec:hadamard}
While microscaling reduces the dynamic range needed to represent tensor values, outliers can still have a disproportionate impact~\citep{an2025systematicoutlierslargelanguage,park2025outliersafepretrainingrobust4bit,raman2025rethinkingoutlierdistributionlarge,dettmers2022llmint88bitmatrixmultiplication,smoothquant} on FP4 formats, degrading model accuracy. Random Hadamard transforms~\citep{shah2024flashattention3fastaccurateattention,ashkboos2025halohadamardassistedlowerprecisionoptimization,quarot,tseng2024quipbetterllmquantization,tseng2025qtipquantizationtrellisesincoherence,malinovskii2024pushinglimitslargelanguage} address this by redistributing outliers into an approximately Gaussian distribution, making them easier to represent in narrower formats. Below we discuss the application of Random Hadamard transforms in FP4 training.

\textbf{GEMMs transformed:}
Random Hadamard transforms are typically applied on both GEMM inputs so that the dot-product inverts each transform by the other operand due to orthogonality. More details on their mechanics is discussed in Appendix~\ref{app:rht_mechanics}. Empirically, we observe that transforming Wgrad inputs improves training for the 12B model (e.g., Figure~\ref{fig:ablations12b} shows that loss worsens after removing transformations from Wgrad). On the other hand, Hadamard transforms show no measurable benefit for Fprop and Dgrad at smaller scales (see Appendix~\ref{app:rht_tensors}), likely because
FP4 already provides sufficient range. As a result, we restrict Hadamard transforms to Wgrad inputs, though there may be cases where Fprop and Dgrad would also benefit.

\textbf{Hadamard matrix size:}
Random Hadamard transforms are implemented as matrix multiplications between \(d\times d\) Hadamard matrices and each tile of the tensor of equal size. The matrix size \(d\) introduces a trade-off between accuracy and performance. Larger matrices distribute outliers more effectively, by spreading them over more values, but increase compute and memory costs. Matrices with too few entries are less likely to reproduce a Gaussian distribution, harming FP4 accuracy. At small scales, we observe no measurable differences in convergence due to matrix size. At larger scales, such as the 12B model, we observe diminishing gains from Hadamard matrices beyond moderate sizes (see Appendix~\ref{app:rht_dimension}), whereas having too few matrix entries affects convergence. We believe this is in part due to larger models having more outliers. We therefore choose a matrix size of $d=16$, which we find to have better convergence than $d=4$ and similar results as $d=128$.

\textbf{Random sign vector:}
Random Hadamard transforms introduce randomness by multiplying with a random diagonal sign vector that flips the signs for entire rows or columns. This reduces the chance that ``structured'' outliers (e.g., tensor patterns aligned with the Hadamard basis) survive the transform. At small scales, randomization has no impact on accuracy, and training remains stable with the standard Hadamard transform. However, we find that randomization benefits larger models trained over longer token horizons, as detailed in Appendix~\ref{app:rht_seed}. In our setup, we use a single random sign vector that is shared across all linear layers throughout training. Our studies show no measurable impact from increasing the number of random sign vectors.

\subsection{2D scaling}\label{sec:2dscaling}
During training, transform and scaling operations apply along 
the dot-product dimension, causing tensors to be transformed and scaled differently in the forward (along rows) and backward (along columns) passes. This occurs because the backward pass transposes the tensors, which changes the dot-product dimension. As a result, the same tensor can have two distinct quantized representations, effectively breaking the chain rule since backpropagation no longer differentiates the same function used in the forward pass. More precisely, the backward update \(\partial x=w_\mathrm{bprop}^T \partial y\) computes a gradient for a different function \(y_\mathrm{bprop}=w_\mathrm{bprop}x\) than used in the forward pass \(y_\mathrm{fprop}=w_\mathrm{fprop}x\) when \(w_\mathrm{fprop} \neq w_\mathrm{bprop}\). 
We hypothesize that chain rule violations in the weights contribute to reduced model accuracy.

\textbf{Block scaling:}
To mitigate this issue, we propose a two-dimensional (2D) block scaling method that ensures consistent quantization in both forward and backward passes. For weights, elements are grouped and scaled in \(16\times16\) blocks (i.e., 16 input channels by 16 output channels) similar to~\cite{liu2024deepseekv3}. 2D block scales are replicated for each of the $1\times16$ blocks when being passed into Tensor Cores, and continue to leverage an FP32 per-tensor scale. Activations and gradients use the standard NVFP4 scaling (i.e., \(1\times16\) blocks), since finer-grained scaling improves quantization accuracy. While activation quantization also presents a chain rule concern, we observe that training is less sensitive to inconsistencies in activation tensors than weight tensors (Appendix~\ref{app:2d} discusses this further). Weights are also more tolerant to the scale granularity because they can adapt to the FP4 values. As illustrated in Figure~\ref{fig:ablations12b}, maintaining consistent quantized weights leads to improved training loss for the 12B model.

\textbf{Random Hadamard transforms:}
Similar to scaling, Random Hadamard transforms applied along the dot-product dimension introduce inconsistency after quantization (i.e., different transformations will result in different quantized values) and, therefore, are not applied on the weight tensors. As a result, transformed activations and gradients in weight-related GEMMs can no longer be inverted by transforming the weight tensor, preventing Fprop and Dgrad from benefiting from the transformation. Therefore, we restrict Hadamard transforms to the Wgrad tensors, which we find sufficient for training our models (Appendix~\ref{app:rht_tensors}).

\subsection{Stochastic rounding}\label{sec:sr}
During quantization to FP4, deterministic rounding (e.g., round-to-nearest-even) can introduce bias, producing systematic errors due to mantissa distributions that favor rounding in a particular direction, values underflowing to zero, or values saturating to the largest representable number.
The effect of bias is typically more pronounced in gradient tensors~\citep{quartet,trainingllmswithmxfp4,fp4alltheway,chmiel2023accurate,alistarh2017qsgd}, which can impact training convergence. To address this bias, we adopt stochastic rounding during quantization of high precision values to FP4. 
Stochastic rounding rounds a value probabilistically to one of its two nearest representable numbers, with probabilities inversely proportional to their distances.
This prevents values from being consistently quantized in the same direction, thereby reducing bias.

We observe that applying stochastic rounding to gradient tensors is essential for convergence in the 12B model, as illustrated in Figure~\ref{fig:ablations12b}. Other tensors in the backward pass do not benefit from stochastic rounding, reinforcing that gradients are the primary source of bias (see Appendix~\ref{app:sr}). Moreover, applying stochastic rounding to the forward pass tensors is detrimental, as it amplifies quantization error relative to nearest rounding~\citep{quartet}.

%%%%%%%%%%%%%%%%%%%%%%%%%%%%%%%%%
\section{NVFP4 and MXFP4}\label{sec:mxfp4}
As discussed earlier, there are two FP4 microscaling formats on NVIDIA Blackwell -- MXFP4 and NVFP4. In this section, we compare training behavior when using these two formats.

\begin{figure}[htbp]
  \centering
  \begin{subfigure}[b]{0.49\textwidth}
    \centering
    \includegraphics[width=\linewidth]{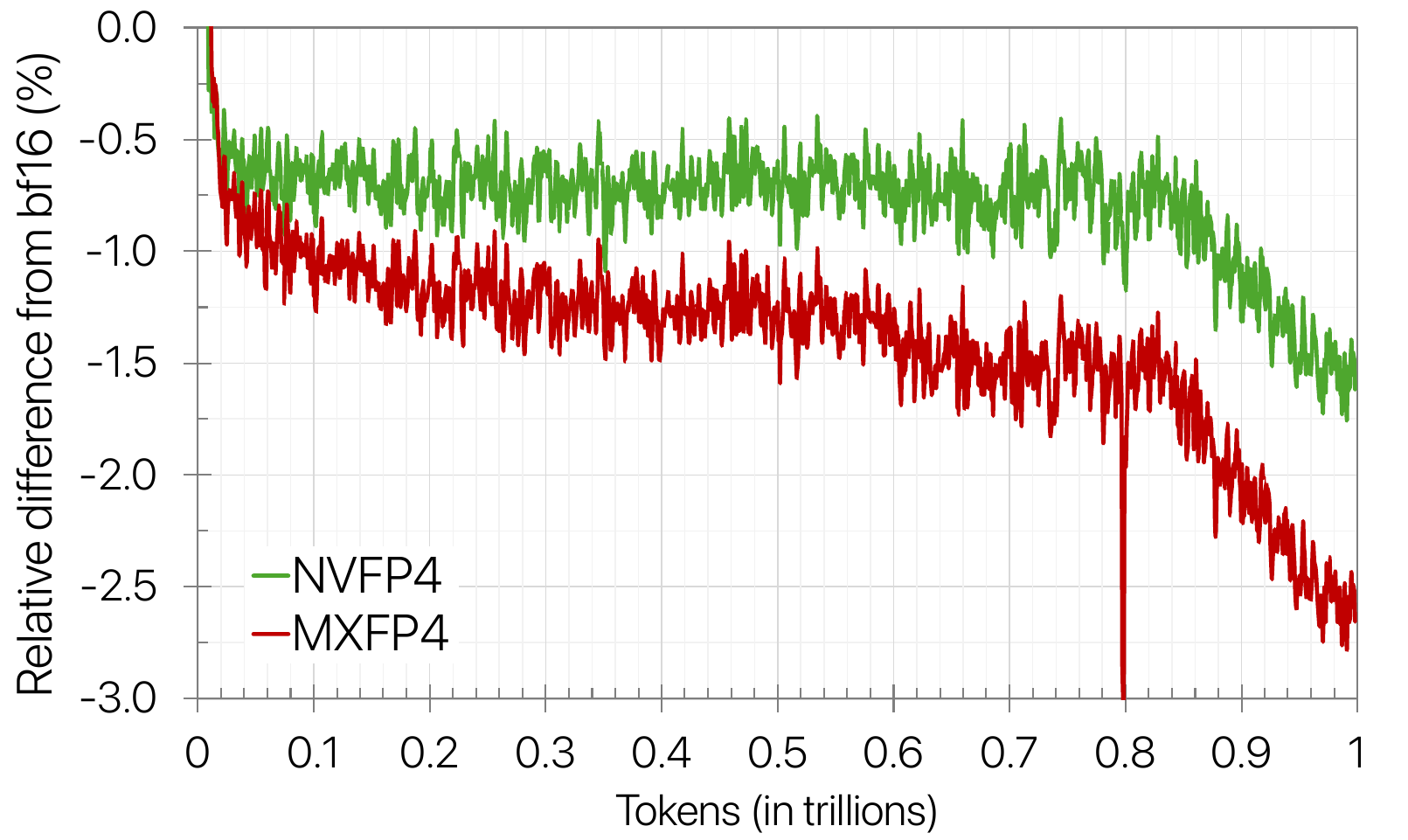}
    \caption{Relative difference between training loss of BF16 (baseline) and NVFP4 and MXFP4 pretraining.}
    \label{fig:nvp4mxfp4loss}
  \end{subfigure}\hfill
  \begin{subfigure}[b]{0.49\textwidth}
    \centering
    \includegraphics[width=\linewidth]{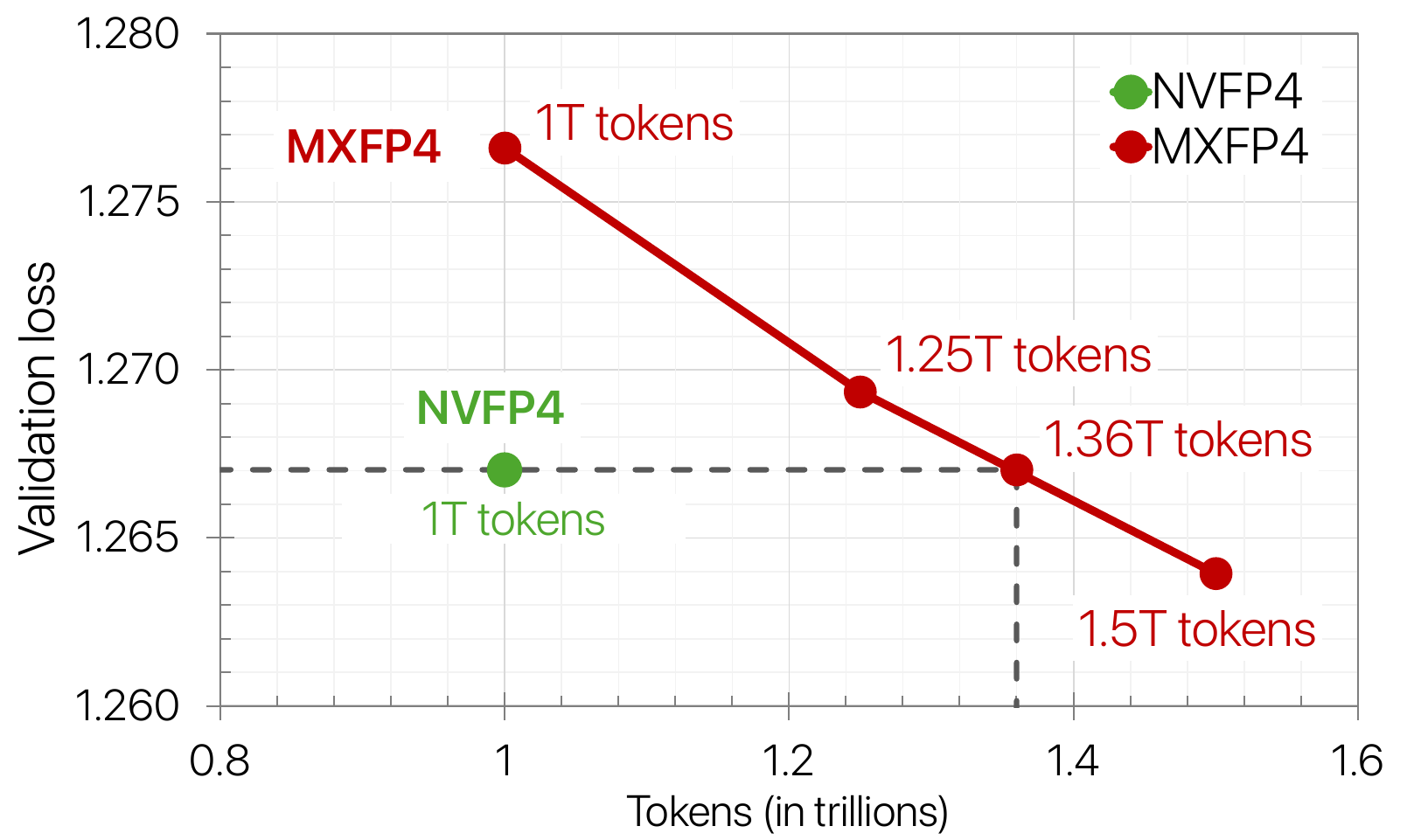}
    \caption{Final validation loss for NVFP4 and MXFP4 pretraining with different number of tokens.}
    \label{fig:tokenpareto}
  \end{subfigure}
  \caption{NVFP4 vs MXFP4 comparisons: (a) training-loss difference; (b) validation perplexity across token budgets.}
  \label{fig:mx_nv_combined}
\end{figure}

\textbf{Model and training setup:}\label{sec:modeldata8b}
We consider an 8-billion parameter (8B) model based on the hybrid Mamba-Transformer architecture. The model is trained on 1 trillion tokens with the same dataset as used for the 12B model. Training consists of two phases of data-blending, between the first $60\%$ and last $40\%$ of training. The model and training details are described in Appendix~\ref{8bmodelarch}.

The reference model is pretrained in BF16. FP4 pretraining follows the training methodology described in Section~\ref{sec:recipe} with MXFP4 and NVFP4 as the respective data formats. For MXFP4, we adopt a Random Hadamard transform size of $d=32$ for Wgrad inputs, to align with the MXFP4 block size. In both the settings, the last eight blocks (either FFNs or Mamba-2) are kept in BF16, comprising about $15\%$ of the model.

\textbf{Results:}\label{sec:results8b}
Figure~\ref{fig:nvp4mxfp4loss} demonstrates that NVFP4 pretraining converges to a better loss than MXFP4. Specifically, MXFP4 has a relative error of around $2.5\%$ compared to $1.5\%$ for NVFP4. 
To close the gap with NVFP4, we extend MXFP4 pretraining with additional tokens (varying between 1T and 1.5T total tokens). Figure~\ref{fig:tokenpareto} illustrates the final loss obtained as a function of number of tokens used during pretraining. We observe that MXFP4 matches NVFP4 loss when trained on $36\%$ more tokens (i.e., using $1.36$T instead of 1T tokens). 
This translates to a considerable increase in training time for MXFP4, highlighting the benefits of NVFP4. Future studies should evaluate scaling laws for these formats on different parameter counts and token horizons.

%%%%%%%%%%%%%%%%%%%%%%%%%%%%%%%%%
\section{Conclusions}

We have demonstrated that large-scale pretraining with NVFP4 is both stable and accurate when paired with a targeted methodology designed to improve training stability and convergence through techniques such as 2D weight scaling, Random Hadamard transforms, stochastic rounding, and others described in this technical report. Using this approach, a 12B hybrid Mamba-Transformer model was trained on 10 trillion tokens, with loss and downstream accuracy closely tracking the FP8 baseline. This establishes the first public evidence of sustained 4-bit pretraining at multi-trillion-token scale.

In side-by-side experiments, NVFP4 reached comparable loss with fewer tokens than MXFP4, indicating efficiency gains without sacrificing accuracy. These comparisons provide an initial view into the memory and compute efficiency benefits, as well as the convergence trade-offs, of different FP4 formats during pretraining.

Future work will further characterize NVFP4’s pretraining performance relative to other formats, while refining the methodology to quantize all linear layers without impacting convergence, reducing remaining high-precision layers, and extending NVFP4 to attention and communication paths. We also plan to explore its use in post-training scenarios and evaluate it on larger models, longer token horizons, and additional architectures such as mixture-of-experts. NVFP4 training on Blackwell is now fully supported via a recent update to \href{https://github.com/NVIDIA/TransformerEngine/pull/2177}{Transformer Engine}.

%%%%%%%%%%%%%%%%%%%%%%%%%%%%%%%%%%%%%%%%%%%%%%%%%%%%%%
\newpage

%%%%%%%%%%%%%%%%%%%%%%%%%%%%%%%%%
\section*{Contributors}

\textbf{Numerics, Evaluations}: Anjulie Agrusa, Muya Chang, Mike Chrzanowski, Eric Chung, Steve Dai, Bita Darvish Rouhani, Carlo del Mundo, Brucek Khailany, Mikail Khona, Nick Knight, Ben Lanir, Simon Layton, Daniel Lo, Paulius Micikevicius, Asit Mishra, Deepak Narayanan, Chao Ni, Mostofa Patwary, Sweta Priyadarshi, Yigong Qin, Oleg Rybakov, Charbel Sakr, Sanjeev Satheesh, Mohammad Shoeybi, Michael Siu, Darko Stosic, Dusan Stosic, Bor-Yiing Su, Nima Tajbakhsh, Aditya Vavre, Rangharajan Venkatesan, Roger Waleffe, Qiyu Wan, Mengdi Wang, Lizzie Wei, Hao Wu, Keith Wyss, Jinze Xue

\textbf{SW Support, Performance:} Felix Abecassis, Anjulie Agrusa, Michael Andersch, Jinhang Choi, Victor Cui, Carlo del Mundo, Burc Eryilmaz, Abhinav Goel, Oleg Goncharov, Robert Hesse, Herbert Hum, Ronny Krashinsky, Tim Moon, Yigong Qin, Xiaowei Ren, Kirthi Shankar, Frank Sun, Przemek Tredak, Evgeny Tsykunov, Qiyu Wan, Lizzie Wei, Evan Wu, Keith Wyss, Jinze Xue, Charlene Yang, Yujia Zhai, Jingyang Zhu, Zhongbo Zhu

\textbf{Infrastructure:} Dong Ahn, Stefania Alborghetti, Sivakumar Arayandi, Alexis Bjorlin, Aaron Blakeman, Evan Briones, Carlo del Mundo, Deena Donia, Henry Estela, Yugi Guvvala, Russell J. Hewett, Alex Kondratenko, Deepak Narayanan, Abhijit Paithankar, Satish Pasumarthi, Ankit Patel, Ashwin Poojary, Gargi Prasad, Oleg Rybakov, Stas Sergienko, Pasha Shamis, Nishant Sharma, Misha Smelyanskiy, Shelby Thomas, Evgeny Tsykunov, Gandhi Vaithilingam, Roger Waleffe, Hexin Wang, Ning Xu, Ruoxi Zhang

\textbf{Leadership:} Jonah Alben, Ian Buck, Bryan Catanzaro, Eric Chung, Ujval Kapasi, Michael Lightstone, Mohammad Shoeybi

\newpage
\begin{spacing}{0.9}
\bibliography{references}
\bibliographystyle{references}
\end{spacing}

\newpage
\appendix

%%%%%%%%%%%%%%%%%%%%%%%%%%%%%%%%%
\section*{Appendix}

%%%%%%%%%%%%%%%%%%%%%%%%%%%%%%%%%
\section{Models}\label{app:models}

We evaluate three model variants throughout this technical report: two hybrid Mamba-Transformer architectures at 12B and 8B scales, and a Transformer variant at 1.2B scale. %a two Transformer variants at 8B and 1.2B scales. 
The 12B model is used as the primary architecture to validate NVFP4 training method while the 8B hybrid model is used to compare NVFP4 against MXFP4.
The 1.2B model is used for several ablation studies. 
This section describes the architectural details, datasets, and training schedules used for each model.

\subsection{12B hybrid Mamba-Transformer}\label{12bmodelarch}
\textbf{Model architecture:} Table~\ref{tab:modelarch12b} summarizes the configuration for the 12B hybrid Mamba-Transformer architecture. 
The model has 62 blocks with 6 Self-Attention, 28 FFNs, and 28 Mamba-2 blocks (each block has 2 linear layers). Mamba-2 blocks have 8 groups, state dimension of 128, head dimension of 64, expansion factor of 2, and convolution window size of 4. Squared ReLU activations are used for FFN blocks, RMSNorm~\citep{zhang2019rmsnorm} for the normalization layers, and separate embedding and output layer weights. The model does not have any position embeddings, dropout, or biases for linear layers. Residual skip connections are added to each block.

\begin{table}[htb]
\centering
\caption{Summary of the 12B Nemotron-H hybrid Mamba–Transformer architecture.}
\label{tab:modelarch12b}
\small
\begin{tabular}{ccccccc}
\toprule
\textbf{Number of} & \textbf{Model} & \textbf{FFN} & \textbf{Q} & \textbf{KV} & \textbf{State} & \textbf{Mamba} \\
\textbf{blocks} & \textbf{dimension} & \textbf{dimension} & \textbf{heads} & \textbf{heads} & \textbf{dimension} & \textbf{groups}  \\
\midrule
62 & 5120 & 20480 & 40 & 8 & 128 & 8 \\
\bottomrule
\end{tabular}
\end{table}

\textbf{Dataset:} For the pretraining data, we use a corpus of high-quality curated and synthetic dataset comprising of 10 trillion tokens based on~\cite{nemotronnano2}, with data mixtures consisting of general web crawl data, wikipedia, math, code, academic data, crawl++, multilingual, and synthetic SFT-style data. Pretraining uses a phased data-blending approach~\citep{feng2024twophase}, where the first phase covers 70\% of training with a data mixture that promotes diversity in the data, while the second and third phases primarily consist of high-quality datasets and span the last 20\% and 10\% of training, respectively.

\textbf{Hyperparameters:} The model is trained on 10 trillion tokens using a sequence length of \(8192\) and batch size of \(736\). The WSD schedule has a constant learning rate of $4.5 \cdot 10^{-4}$ that decays to $4.5 \cdot 10^{-6}$ over the last 20\% of training. Adam parameters are \(\beta_1 = 0.9\) and \(\beta_2 = 0.95\), and weight decay is set to \(0.1\). 

\textbf{Precisions:} The reference model is trained in FP8 following the methodology in~\cite{nemotronnano2}. Specifically, all linear layers are computed in E4M3, except the linear layers in the first block and the last two blocks which are left in BF16. Scale factors apply on 128$\times$128 blocks for weights and 1$\times$128 blocks for activations and gradients. They are computed online for each block, stored in FP32, and applied before quantizing the tensor into the FP8 format. Precisions for other operations are the same as in Section~\ref{sec:bf16layers}.

For NVFP4, we follow the method described in Section~\ref{sec:recipe}. All linear layers are computed in NVFP4, except the linear layers in the first two blocks and the last eight blocks (FFNs or Mamba-2) which are left in BF16. This accounts for $16\%$ of the total linear layers kept in high precision.

\subsection{8B hybrid Mamba-Transformer}\label{8bmodelarch}

\textbf{Model architecture:} The 8B hybrid Mamba-Transformer has a similar architecture as the 12B hybrid model. Table~\ref{tab:modelarch8bhybrid} summarizes the configuration for the 8B model. This model has 52 blocks: 4 Self-Attention, 24 FFNs, and 24 Mamba-2 blocks. The model hidden dimension is 4096, FFN hidden dimension is 21504, and Grouped-Query Attention has 32 query heads along with 4 key-value heads. Mamba-2 blocks have 8 groups, state dimension of 128, head dimension of 64, expansion factor of 2, and convolution window size of 4.

\begin{table}[htb]
\centering
\caption{Summary of the 8B Nemotron-H hybrid Mamba–Transformer architecture.}
\label{tab:modelarch8bhybrid}
\small
\begin{tabular}{ccccccc}
\toprule
\textbf{Number of} & \textbf{Model} & \textbf{FFN} & \textbf{Q} & \textbf{KV} & \textbf{State} & \textbf{Mamba} \\
\textbf{blocks} & \textbf{dimension} & \textbf{dimension} & \textbf{heads} & \textbf{heads} & \textbf{dimension} & \textbf{groups}  \\
\midrule
52 & 4096 & 21504 & 32 & 4 & 128 & 8 \\
\bottomrule
\end{tabular}
\end{table}

\textbf{Hyperparameters:} The model is trained on 1 trillion tokens from the same dataset used for the 12B model. A batch size of $768$ is used with only two phases of data-blending, split between the first $60\%$ and last $40\%$ of training. The sequence length is \(8192\) and the WSD schedule uses a constant learning rate of $8.0 \cdot 10^{-4}$ that decays to $8.0 \cdot 10^{-6}$ over the last 15\% of training. Adam parameters are \(\beta_1 = 0.9\) and \(\beta_2 = 0.95\), and weight decay is set to \(0.1\). 

\textbf{Precisions:}
The reference model is trained in BF16. For NVFP4, we follow the methodology described in Section~\ref{sec:recipe}. All linear layers are computed in NVFP4, except for the linear layers in the last eight blocks (FFNs or Mamba-2) which are left in BF16.

\subsection{1.2B Transformer}\label{1.2bmodelarch}

\textbf{Model architecture:} The 1.2B model follows the standard Transformer architecture. Details on the model configuration are summarized in Table~\ref{tab:modelarch1.2b}. The model has $20$ transformer blocks, each comprising of Self-Attention and FFN blocks. The model hidden dimension is 2048, FFN hidden dimension is 6144, and Self-Attention has 16 query heads and 8 key and value heads. FFN blocks use squared ReLU activations. The model uses RoPE embeddings and does not have any dropout or biases for linear layers. Residual skip connections are added to each of the transformer blocks.

\begin{table}[ht]
\centering
\caption{Summary of the 1.2B Nemotron Transformer architecture.}
\label{tab:modelarch1.2b}
\small
\begin{tabular}{cccccccc}
\toprule
\textbf{Number of} & \textbf{Model} & \textbf{FFN} & \textbf{Head} & \textbf{Q} & \textbf{KV} \\
\textbf{blocks} & \textbf{dimension} & \textbf{dimension} & \textbf{dimension} & \textbf{heads} & \textbf{heads} \\
\midrule
20 & 2048 & 6144 & 128 & 16 & 8  \\
\bottomrule
\end{tabular}
\end{table}

\textbf{Hyperparameters:} The model is trained on 1 trillion tokens with the same dataset as for the 8B model, using two phases of data-blending. The model is trained with a sequence length of \(8192\) and batch size of \(768\). The WSD schedule starts with a learning rate of $1.2\cdot10^{-3}$ for $85\%$ of training that decays to $1.2\cdot10^{-5}$ for the last $15\%$ of training.

\textbf{Precisions:}
The reference model is trained in BF16. For NVFP4, we perform ablations on the methodology from Section~\ref{sec:recipe}. Linear layer tensors are converted to NVFP4. Precisions for other operations are the same as in Section~\ref{sec:bf16layers}.

%%%%%%%%%%%%%%%%%%%%%%%%%%%%%%%%%
\section{NVFP4 Quantization Procedure}\label{app:quantization}
The procedure for converting a tensor from higher precision (FP32 or BF16) to NVFP4 is described below. Given a tensor \(x\), each block \(b\) of contiguous high-precision values \(x_i\), \(i \in b\) is quantized to FP4. Prior to quantization, values are scaled using a two-level scaling strategy: first, a global FP32 tensor-level scale factor moves all the values within a tensor into representable range of a block (FP4 $\times$ FP8); second, a local block-level scale factor moves the values \(x_i\) within a block into FP4 representable range.

\subsection{Global tensor-level scaling}
The global \textit{encode} scale is computed as:
\begin{equation}\label{eqn:encodingscale}
\begin{split}
    s_{enc} &= \frac{6 \cdot 448}{amax_{x}} \\ 
\end{split}
\end{equation}
where \(amax_{x}=\underset{i}{\max}(|x_i|)\) represents the absolute maximum value across the entire tensor $x$, $6$ and $448$ are the maximum representable magnitudes in the E2M1 and E4M3 formats, respectively.
The corresponding \textit{decode} scale, \(s_{dec}=1/s_{enc}\), gets stored in FP32 for decoding the resulting values after the NVFP4 GEMM operation. Since the global scale is computed dynamically across the entire tensor, it induces an extra pass through device memory: once to compute the global amax, and once to scale prior to conversion to FP4, as described later. However, the global scale could potentially span a smaller granularity (e.g., a row or block of elements) to avoid additional round-trips through device memory.

\subsection{Local block-level scaling}
The local \textit{decode} scales are chosen so the largest absolute value in each block, \(amax_b=\underset{i \in b}{\max}(|x_i|)\), normalizes to the FP4 maximum representable:
\begin{equation}
\begin{split}
s_{dec,b} = \frac{amax_b}{6}
\end{split}
\end{equation}

Since the local \textit{decode} scales must be stored in FP8 for Tensor Cores, they are first multiplied by the global \textit{encode} scale before quantization:

\begin{equation}\label{eq:decode_scale}
\begin{split}
s_{dec,b,e4m3}=\mathrm{e4m3}(s_{dec,b} \cdot s_{enc}),
\end{split}
\end{equation}

where the goal of \(s_{enc}\) is to remap the largest local \textit{decode} scale, i.e., \({max}(s_{dec,b})=amax_x/6\), to the FP8 maximum representable. We obtain the real local \textit{encode} scale factor by inverting the quantized local \textit{decode} scale in higher precision and scaling it back to its original representable range, \(s_{enc,b}=1/(\mathrm{fp32}(s_{dec,b,e4m3}) \cdot s_{dec})\). In this way, we try to ensure that the original value can be recovered after scaling with \(s_{enc,b} \cdot s_{dec} \cdot s_{dec,b,e4m3} \approx 1\), since failing to do so can impact model accuracy. Round-to-nearest-even is used when computing the decode scale factor in Equation~\ref{eq:decode_scale}.

\subsection{Conversion}
Combining all of these together, each element \(x_i\) in the block gets scaled by the local \textit{encode} scale and quantized as 
\begin{equation}\label{eqn:quantization_0}
\begin{split}
    \hat{x}_i & = q(x_{i} \cdot s_{enc,b}), \\
\end{split}
\end{equation}
where \(q(\cdot)\) denotes the FP4 quantization function. Beyond storing the quantized values \(\hat{x}_i\), the local and global \textit{decode} scales, \(s_{dec,b,e4m3}\) and \(s_{dec}\), are also stored in memory and used during the matrix multiplication. 

Tensor Core reads the local \textit{decode} scales and applies them to partial dot-products computed over \(b\) elements:
\begin{equation}\label{eqn:quantization_1}
\begin{split}
 s^x_{dec,b,e4m3} \cdot s^y_{dec,b,e4m3} \cdot \sum_{k \in b}(x_k \cdot y_k), \\
\end{split}
\end{equation}
where \(x\) and \(y\) denote the two input operands. After the GEMM operation, the global \textit{decode} scales \(s^x_{dec}\) and \(s^y_{dec}\) are applied on the final output in a similar fashion.

\subsection{Remarks on MXFP4 and NVFP4 scale factor}\label{app:mxfp4scale}
MXFP4 scale factors are restricted to powers-of-two, meaning values can not be scaled to fit perfectly into the FP4 representable range. After scaling, the block amax will either overflow the FP4 maximum representable and saturate, or round down to a smaller FP4 sample. Since saturations have been observed to cause convergence issues for MXFP8 training~\citep{mxfp8recipes}, we typically round decode scale factors up to prevent saturations.

This scaling strategy can result in some FP4 samples being wasted while also reducing the utilized dynamic range. As an example, consider a block of values with an absolute maximum value of \(amax=3+\delta\), where \(\delta\) represents a small increment. In order to move the block amax to the FP4 maximum representable number (i.e., \(\pm6\) for E2M1), the decode scale factor is computed as \(s_{dec,b}=amax/6=0.5+\delta/6\), which rounds up to the next power-of-two, to \(s_{dec,b,ue8m0}=1\). After scaling, the block's amax becomes \(amax/s_{dec,b,ue8m0}=3+\delta\), which quantizes to \(3\) in FP4. As a result, in the worst case, FP4 is unable to represent the samples at $\pm4$ and $\pm6$. This also reduces the dynamic range by nearly one binade, where only $\log_2(3/0.5)=2.58$ binades are utilized instead of the full $\log_2(6/0.5)=3.58$ binades, where $0.5$ represents the minimum positive non-zero magnitude in FP4.

NVFP4 overcomes this limitation with a more precise E4M3 block scale, which maps the block amax much closer to the FP4 maximum representable number. This maximizes the FP4 samples utilization and preserves more of the dynamic range of FP4.

%%%%%%%%%%%%%%%%%%%%%%%%%%%%%%%%%
\section{Hadamard Transform Mechanics}\label{app:rht_mechanics}
Random Hadamard transforms applies an orthogonal rotation to the tensor being quantized, i.e., \(x^{\prime }=q(xH\cdot s)\), where \(H\) is the Hadamard matrix, \(q(\cdot)\) is the quantization function, and $s$ is the scale factor computed in the rotated space $xH$. The Hadamard matrix is defined by normalized matrices of the form $H_d=(1/\sqrt{2})H_{2}\otimes H_{d/2}$ with elements constrained to $\pm1$. Given their orthogonal nature, they can be applied to both operands of a matrix multiplication:
\begin{equation}
C=(AH)(H^TB)=AB,
\end{equation}
where the transform from each operand gets inverted within the dot-product by $HH^T=I$.

Random Hadamard transforms introduce randomization into the transformation by left-hand multiplying a $d$-dimensional diagonal random matrix, $S_d$, with a Hadamard matrix, resulting in $H=S_dH_d$, where diagonal entries of $S_d$ are randomly chosen from $\{-1, 1\}$. The entries in $S_d$ will flip the signs for different rows of $H_d$.

We perform Hadamard transforms in a tiled approach by multiplying \(H\), which has \(d\times d\) matrix entries, with an \(m\times k\) tensor, where every \(d\times d\) elements of the tensor are multiplied by \(H\). The transform involves \(mkd\) multiply-adds and \(d^2\) reads for the Hadamard matrix, which is a small cost when \(d\) is much smaller than the tensor dimensions \(m\) or \(k\). In this case, Hadamard transforms can be implemented as batched matrix multiplications, which are limited by memory traffic from reading the input tensor when using Tensor Cores, and can be fused with other layers to reduce round-trips to device memory.

\begin{figure}[ht] 
    \centering
    \includegraphics[width=0.60\textwidth]{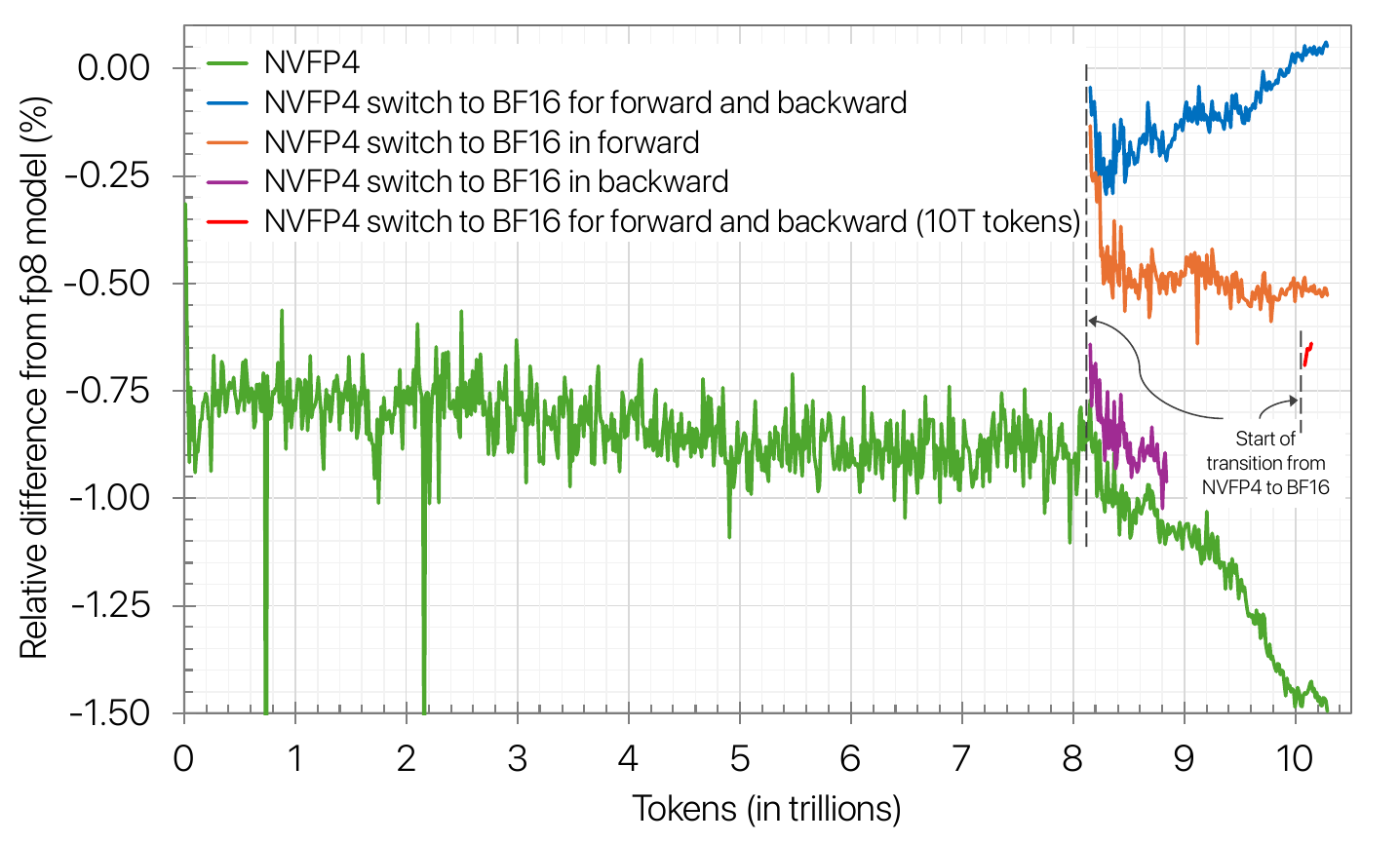}
    \caption{Switching to higher precision towards end of training. Plot shows relative difference in validation loss for a 12B model trained on 10T tokens. NVFP4 uses the method specified in Section~\ref{sec:recipe} during all of the training period (Green). The precision for tensors in forward and backward pass (Blue), tensors only in the forward pass (Orange), and tensors only in the backward pass (Purple) are switched from NVFP4 to BF16 at 8.2T tokens until remainder of training. A run where the switch to high precision occurs around 10T tokens is also shown (Red). 1D weight scaling is used when switching precision for the backward pass, since doing so is marginally better than 2D weight scaling in such a setup.}
    \label{fig:healing12b}
\end{figure}

%%%%%%%%%%%%%%%%%%%%%%%%%%%%%%%%%
\section{Switching to Higher Precision}\label{app:healing}

For situations where FP4 training does not completely match the loss of higher precision training, we observe that switching from FP4 to higher precision towards the end of training can close the loss gap. Figure~\ref{fig:healing12b} shows that loss matches the FP8 baseline when precisions are switched after 8.2T tokens (e.g., for $18\%$ of training) and only slightly worse when switched after 10T tokens (e.g., for less than $1\%$ of training). While switching precisions later in training fails to fully recover, presumably because the learning rate is too low for the weight updates, it significantly reduces the portion of training not performed in FP4. We therefore recommend switching to high precision shortly before the onset of learning rate decay for full loss recovery, or at the very end for notable loss improvements with minimal effect on training runtime.

% We therefore recommend switching to higher precision slightly before learning rate starts to decay for full loss recovery or at the very end for significant improvements in loss without impacting training runtime.

We find that most of FP4 training's loss gap arises from quantizing tensors in the forward pass~\citep{quartet}. 
More specifically, most of the loss in the 12B model is recovered (from $1.5\%$ to $0.5\%$ relative error) by switching to higher precision for the forward pass starting at 8.2T tokens. In contrast to~\cite{fp4alltheway}, which reports loss recovery from switching precision in the backward pass, we observe no such improvement in our models. Focusing on the forward pass minimizes the overhead of switching precision, as only about $6\%$ of the total computations (roughly one-third of the final $18\%$ of training) are performed in higher precision.

%%%%%%%%%%%%%%%%%%%%%%%%%%%%%%%%%
\section{Ablation of Training Methodology}\label{sec:appendix_ablations}

\begin{figure}[bpth] \centering
    \includegraphics[width=0.60\textwidth]{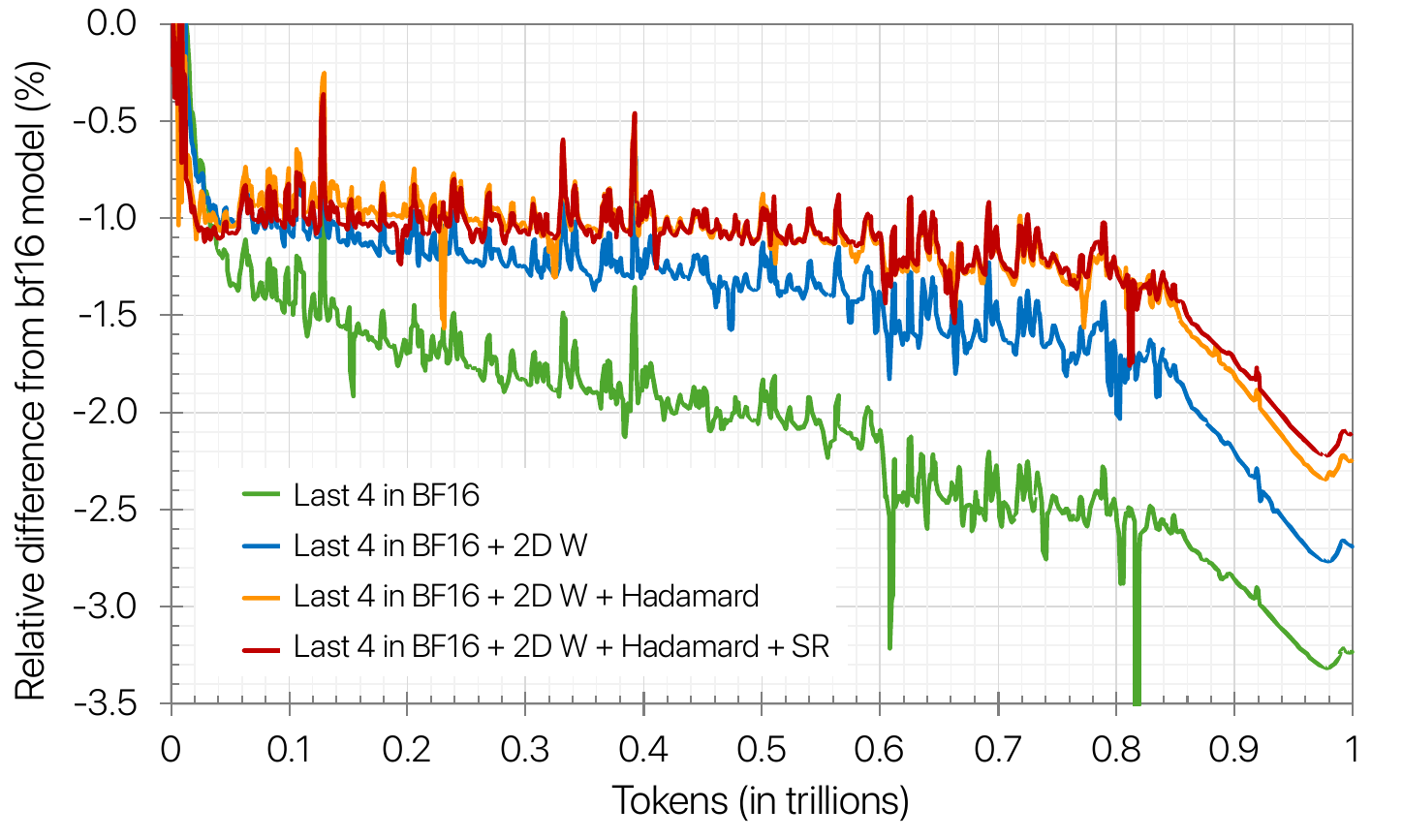}
    \caption{Combining NVFP4 training techniques: linear layers in last four blocks in BF16, 2D weight scaling, Random Hadamard transforms on Wgrad, and stochastic rounding on gradients. Plot shows relative difference in validation loss for a 1.2B model trained on 1T tokens.} 
    \label{fig:recipes}
\end{figure}

\subsection{Combining techniques}\label{app:stacking}
Given FP4 training requires a suite of techniques, we explore the effects of combining them. We start from a base method that quantizes all of the layers to NVFP4, applies the standard NVFP4 scaling (i.e., \(1\times16\) E4M3 per-block scales with FP32 per-tensor scales) to all tensors, and uses round-to-nearest-even on all tensors. This base method is used throughout the appendix and combined with other techniques, unless specified otherwise. Our models diverge early in training when using this base method without any of the additional techniques. We find that maintaining some linear layers in higher precision plays a key role in training stability, as elaborated in the following section. While techniques such as stochastic rounding can improve training stability, they eventually diverge when used in isolation. Figure~\ref{fig:recipes} shows that combining the techniques leads to improvements in the loss. The relative benefit of each technique depends on the order in which the components are added. Combining all of the components together reduces the loss gap compared to a single technique.

\subsection{Layer sensitivity}\label{app:layersensitivity}

\begin{figure}[htbp] \centering
    \includegraphics[
         width=0.60\textwidth,
    ]{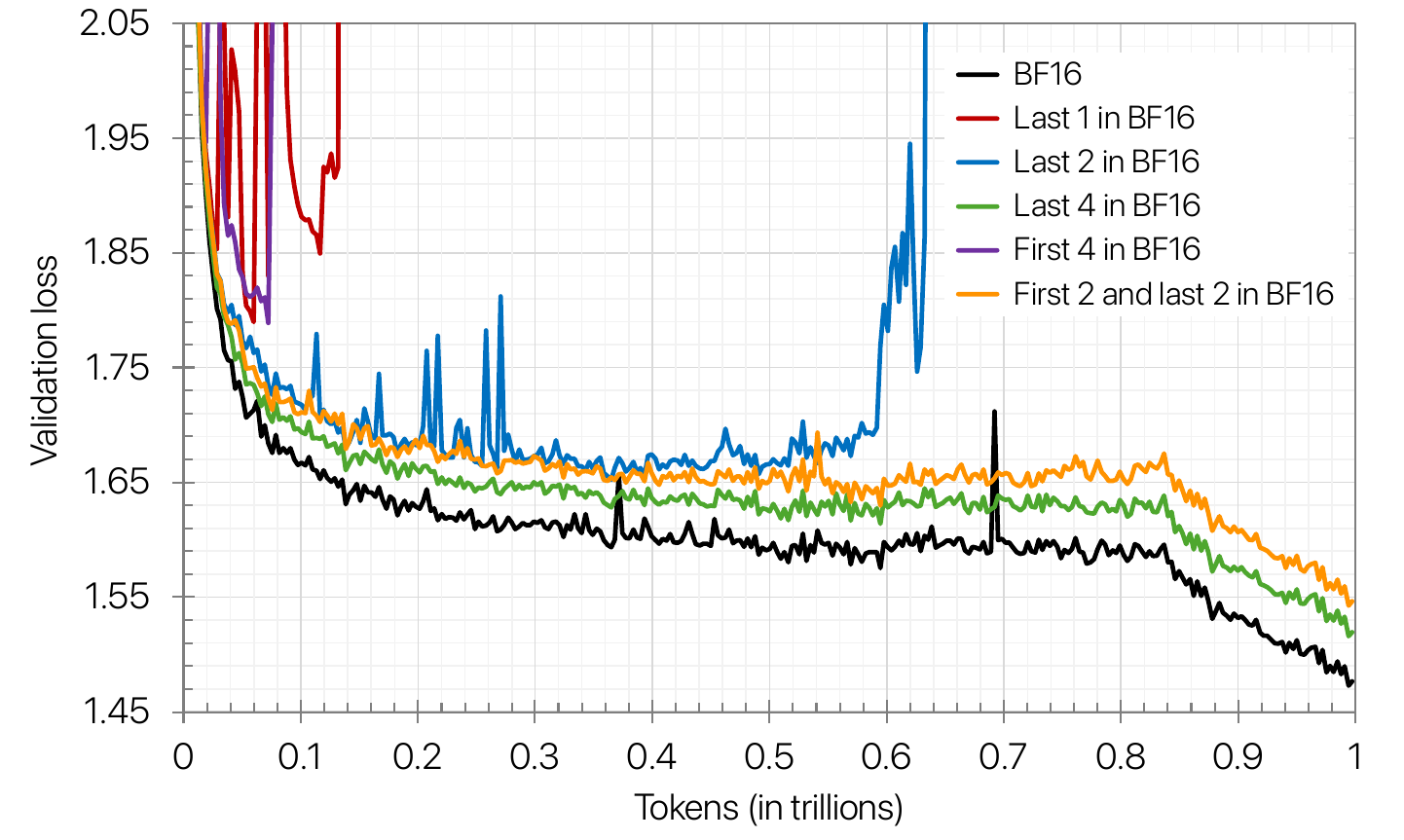}
    \caption{Sensitivity of linear layers to quantization. NVFP4 for all linear layers except in a few of the first and last blocks in the model.
    Plot shows validation loss for a 1.2B model trained on 1T tokens.} 
    \label{fig:bf16_tensors}
\end{figure}

While training diverges with the base method when not using any of the techniques, some layers seem to be more sensitive to FP4 than others. Figure~\ref{fig:bf16_tensors} shows loss converges when the linear layers in the last four blocks remain in BF16, which implies the final layers are more sensitive to FP4 quantization. 
Maintaining the first few blocks in higher precision does not improve stability unless combined with the last blocks (e.g., training is stable when the first two and last two blocks are in BF16, but not when the first four blocks are in high precision).

Based on tensor analysis, we observe the last layers tend to have larger quantization errors in the weight gradients (i.e., Wgrad output from its inputs being FP4). Quantization error metrics could potentially serve as a mechanism to determine which linear layers should remain in higher precision during training.

\subsection{Stochastic rounding on tensors}\label{app:sr}
\begin{figure}[htbp] \centering
    \includegraphics[
         width=0.60\textwidth,
    ]{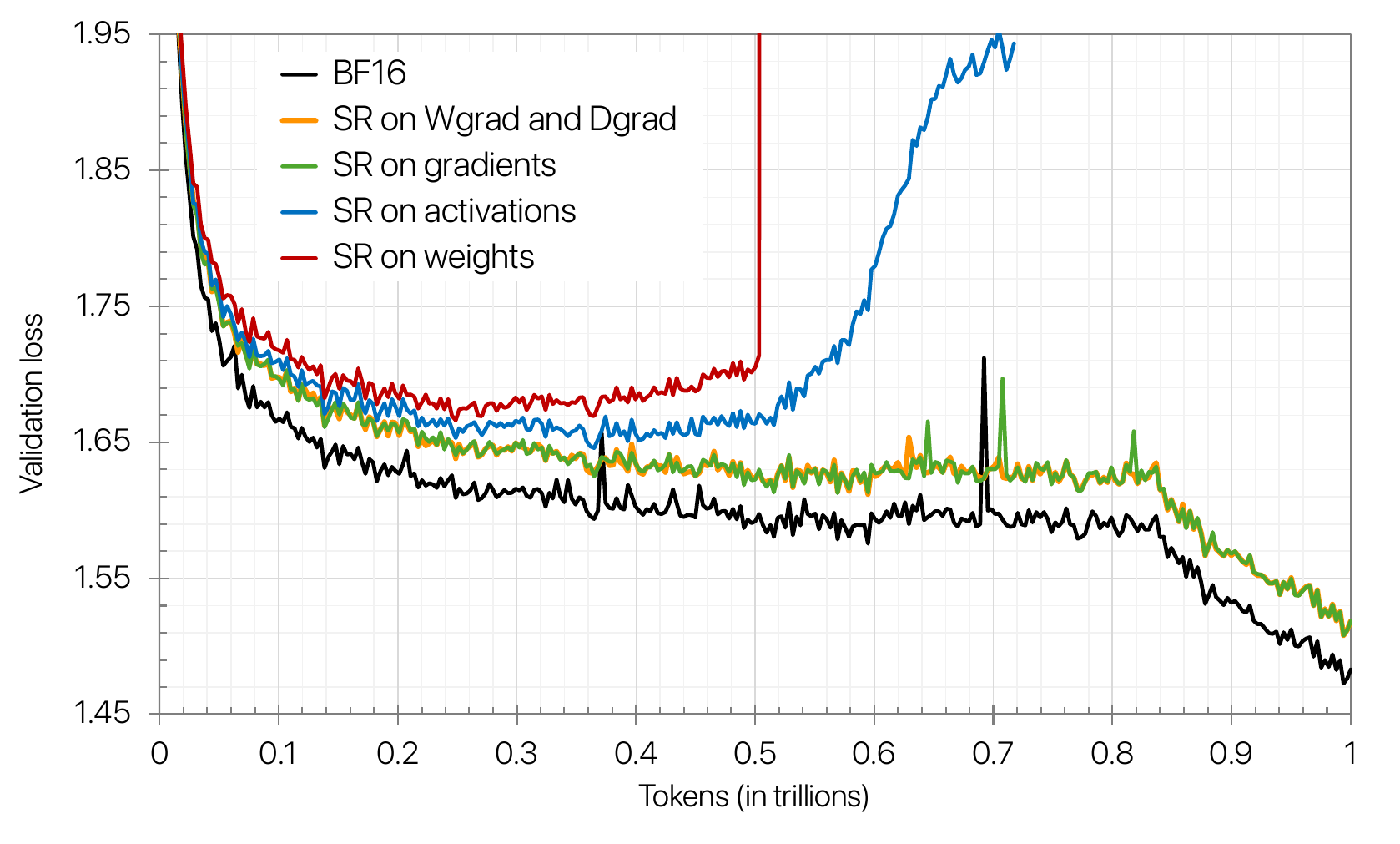}
    \caption{Stochastic rounding applied to different tensors: gradients, activations, weights, and backward-pass tensors. NVFP4 is applied on all linear layers except in the last four blocks. Plot shows validation loss for a 1.2B model trained on 1T tokens.} 
    \label{fig:SR_tensors}
 \end{figure}

Since stochastic rounding is important for FP4 training, we investigate its effect on various tensors during training. As shown in Figure~\ref{fig:SR_tensors}, applying stochastic rounding to gradients leads to stable convergence of the training loss for the 1.2B model, whereas using it on activations or weights causes divergence. 
A potential cause of divergence due to stochastic rounding of activation and weight tensors is that this form of rounding introduces more quantization error than nearest rounding~\citep{fp4alltheway}.
This aligns with prior findings that stochastic rounding mitigates gradient bias arising from quantization~\citep{trainingllmswithmxfp4,fp4alltheway,oscillationreducedmxfp4,quartet}. Additionally, stochastic rounding of all tensors in the backward pass shows little improvement over stochastic rounding of gradients only. This suggests that divergence arises from stochastically rounding tensors in the forward pass. 
For the 12B model, we observe that stochastic rounding must be applied to gradients going into both Dgrad and Wgrad to achieve proper convergence.

\subsection{Random Hadamard transforms}\label{app:rht}

\begin{figure}[htbp] \centering
    \includegraphics[
         width=0.60\textwidth,
    ]{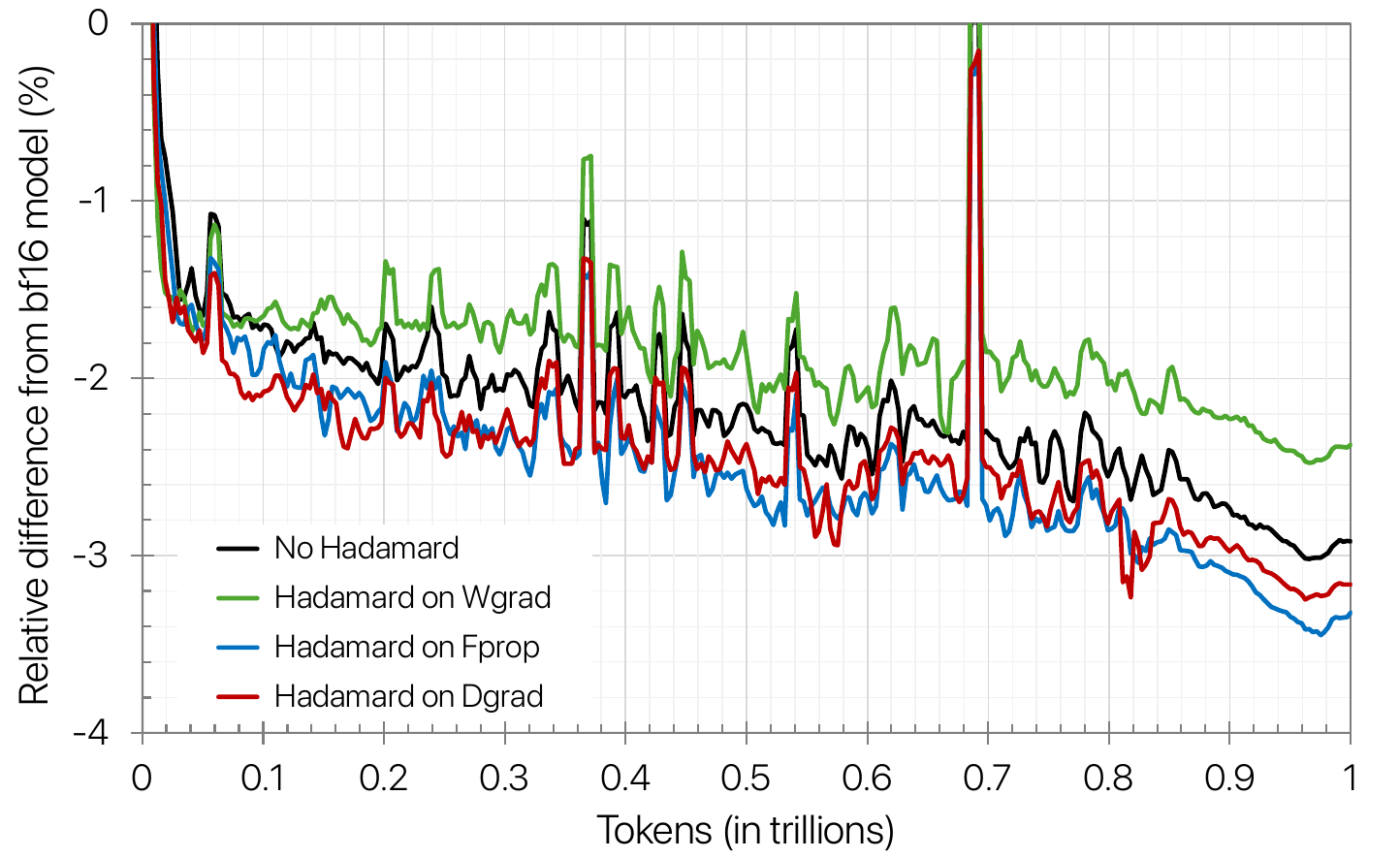}
    \caption{Impact of applying Random Hadamard Transforms (RHT) to different GEMMs (Fprop, Dgrad and Wgrad) during training, compared to no RHT. For RHT runs, each transform uses a fixed random seed across the entire training. NVFP4 quantization is applied to all linear layers except in the last four blocks. The plot shows the relative change in validation loss compared to the BF16 baseline for a 1.2B-parameter model trained on 1T tokens.}
    \label{fig:RHT_gemms}
 \end{figure}

\subsubsection{GEMMs to apply RHT}\label{app:rht_tensors}
We evaluate the impact of applying Random Hadamard Transforms (RHT) to different GEMMs (Fprop, Dgrad and Wgrad) during FP4 training. As shown in Figure~\ref{fig:RHT_gemms}, applying RHT to Wgrad inputs improves validation loss for the 1.2B model, while transforming Fprop or Dgrad inputs degrades model quality. We hypothesize that RHT introduces additional quantization error that offsets the benefit of outlier removal. Thus, although RHT reduces the dynamic range required to represent outliers, its application can negatively affect training when used on certain GEMMs.

\begin{figure}[htb] \centering
    \includegraphics[
         width=0.60\textwidth,
    ]{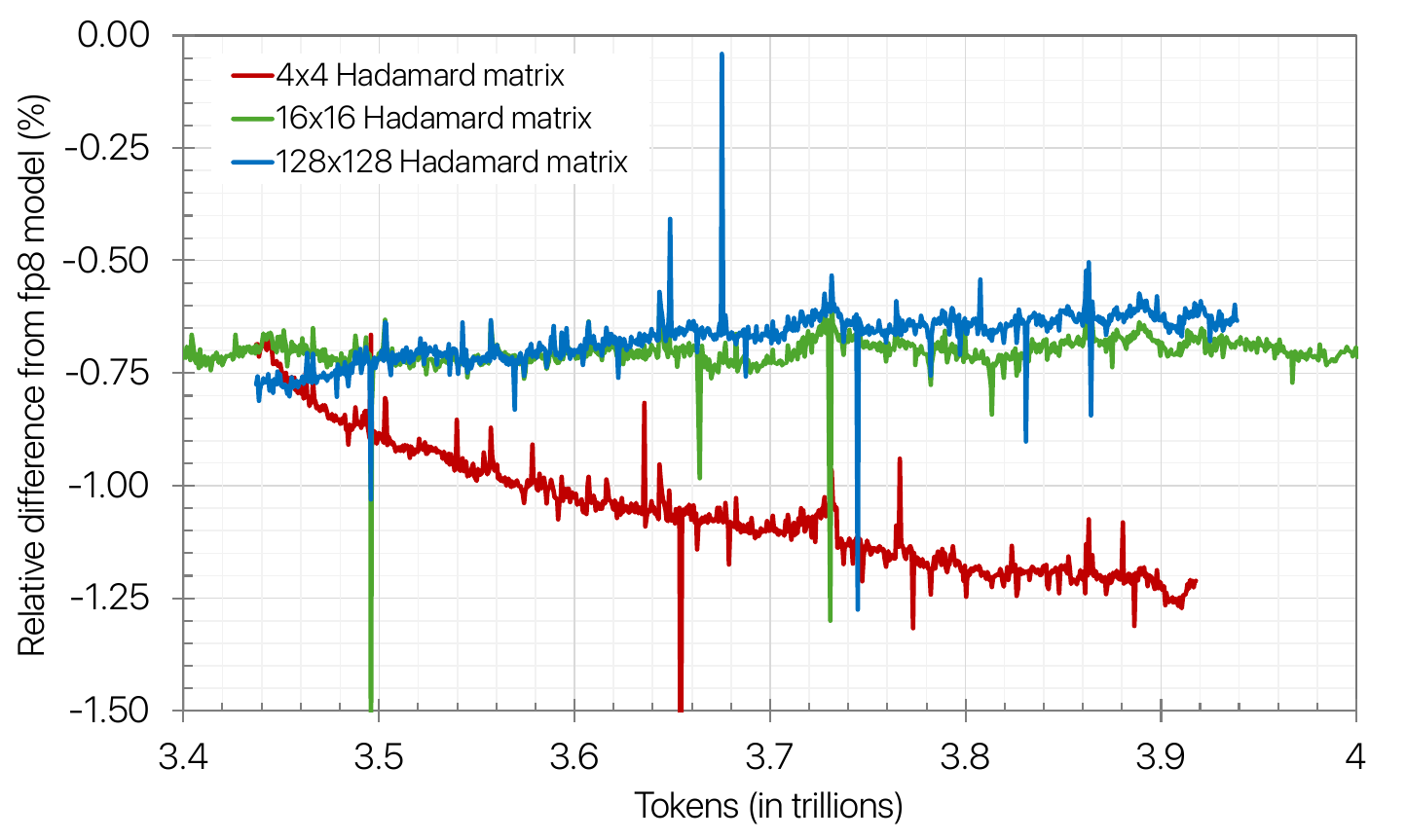}
    \caption{Effect of varying Hadamard Matrix Size. Wgrad tensors use $16\times16$ transforms for the first 3.4T tokens, then switch to $4\times4$ or $128\times128$ for the remainder of training. Plot shows relative difference in training loss for the 12B model trained on 4T tokens. NVFP4 is applied on linear layers using the methodology specified in Section~\ref{sec:recipe}.}
    \label{fig:RHT_dimension}
 \end{figure}

\subsubsection{Hadamard matrix size}\label{app:rht_dimension}

Since the Hadamard matrix size impacts the extent of outlier mitigation, we consider different choices of matrix sizes to transform Wgrad inputs. For the 1.2B model, we observe virtually no difference in loss between $2\times2$, $4\times4$, $16\times16$ and $128\times128$ matrices. To validate this trend at scale, we take the 12B model trained up to 3.4T tokens, switch the matrix size from $16\times16$ to $4\times4$ or $128\times128$, and continue training. 

Figure~\ref{fig:RHT_dimension} shows that $4\times4$ matrices induce an increase in loss and $128\times128$ matrices result in a minor benefit to model quality. This follows the intuition that larger Hadamard matrices can better distribute outliers, whereas matrices with too few entries are less likely to reproduce a Gaussian distribution. The results validate our choice of a $16\times16$ matrix, which reduces the cost of the transform without compromising model accuracy. It also highlights the need to experiment with larger models trained on longer token horizons, since conclusions from smaller scales may not always hold for larger models.

\subsubsection{Role of randomization}\label{app:rht_seed}

\begin{figure}[htb] \centering
    \includegraphics[
         width=0.60\textwidth,
    ]{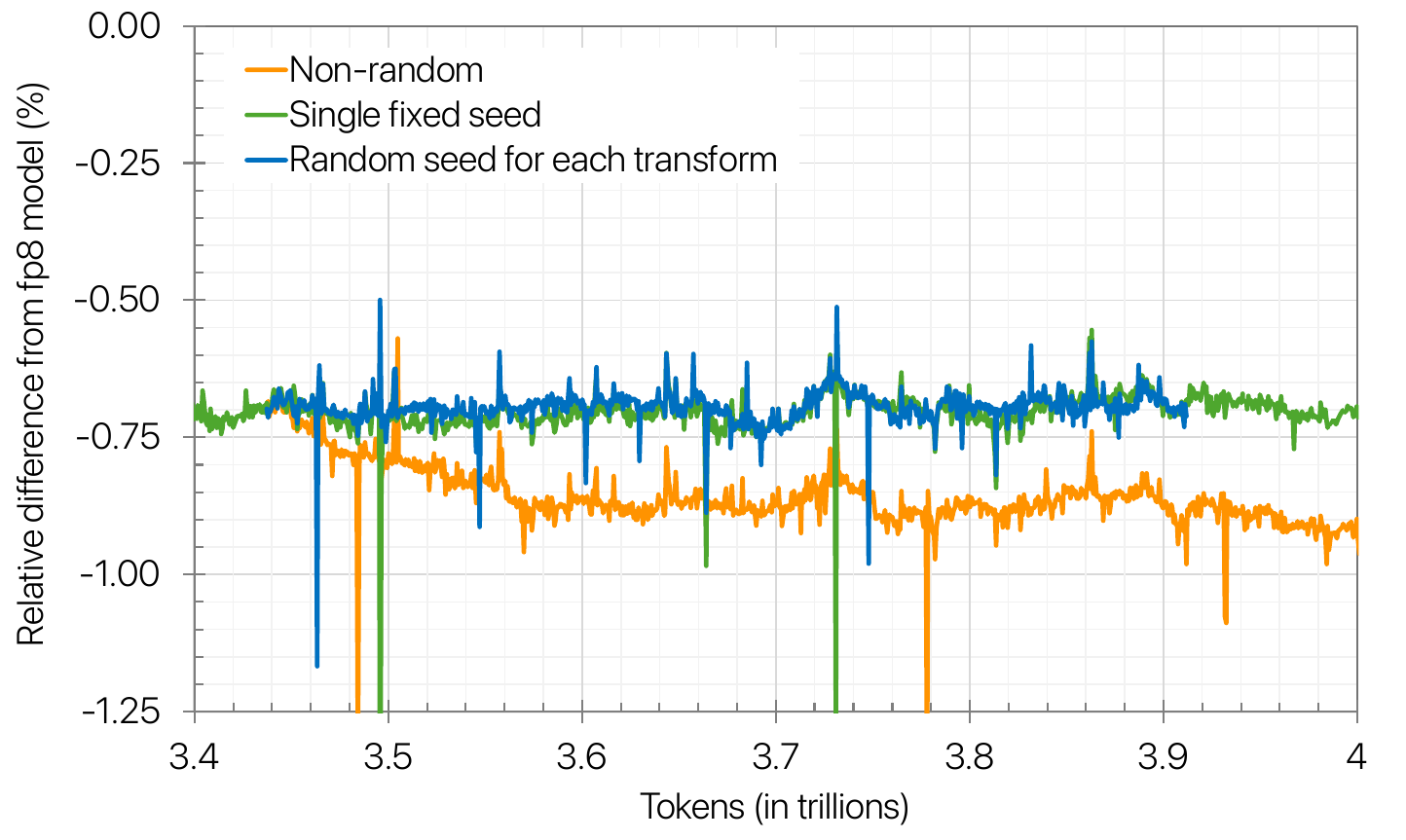}
    \caption{Effect of randomization for the Hadamard transform. A single fixed seed is used for all transforms during the first 3.4 tokens and switched to one of the following randomization options for the remainder of training: a single fixed seed for all layers, a unique seed for every transform, and not using a random sign vector. Plot shows relative difference in training loss from the FP8 baseline for a 12B model trained on 4T tokens. NVFP4 training uses the training methodology specified in Section~\ref{sec:recipe}.} 
    \label{fig:RHT_seed}
\end{figure}

Random Hadamard transforms introduce randomness into the transformation, so we study the importance of this randomization during training. 
Figure~\ref{fig:RHT_seed} illustrates loss when training using different degrees of randomization: (1) ``seed per instance,'' a new random sign vector for every transformation, (2) ``single fixed seed,'' a single random sign vector used for all transformations during training, and (3) no random sign vector. We observe lower model quality in the absence of random sign vectors and no improvements from inducing randomness at every transform instance. A a result, we find it sufficient to use a single fixed seed for all transforms for our 12B model. Interestingly, there are no noticeable differences in model quality between the randomization strategies on the 1.2B model, further confirming that techniques become more critical at larger models and longer token horizons.

\begin{figure}[hbt!] \centering
    \includegraphics[
         width=0.60\textwidth,
    ]{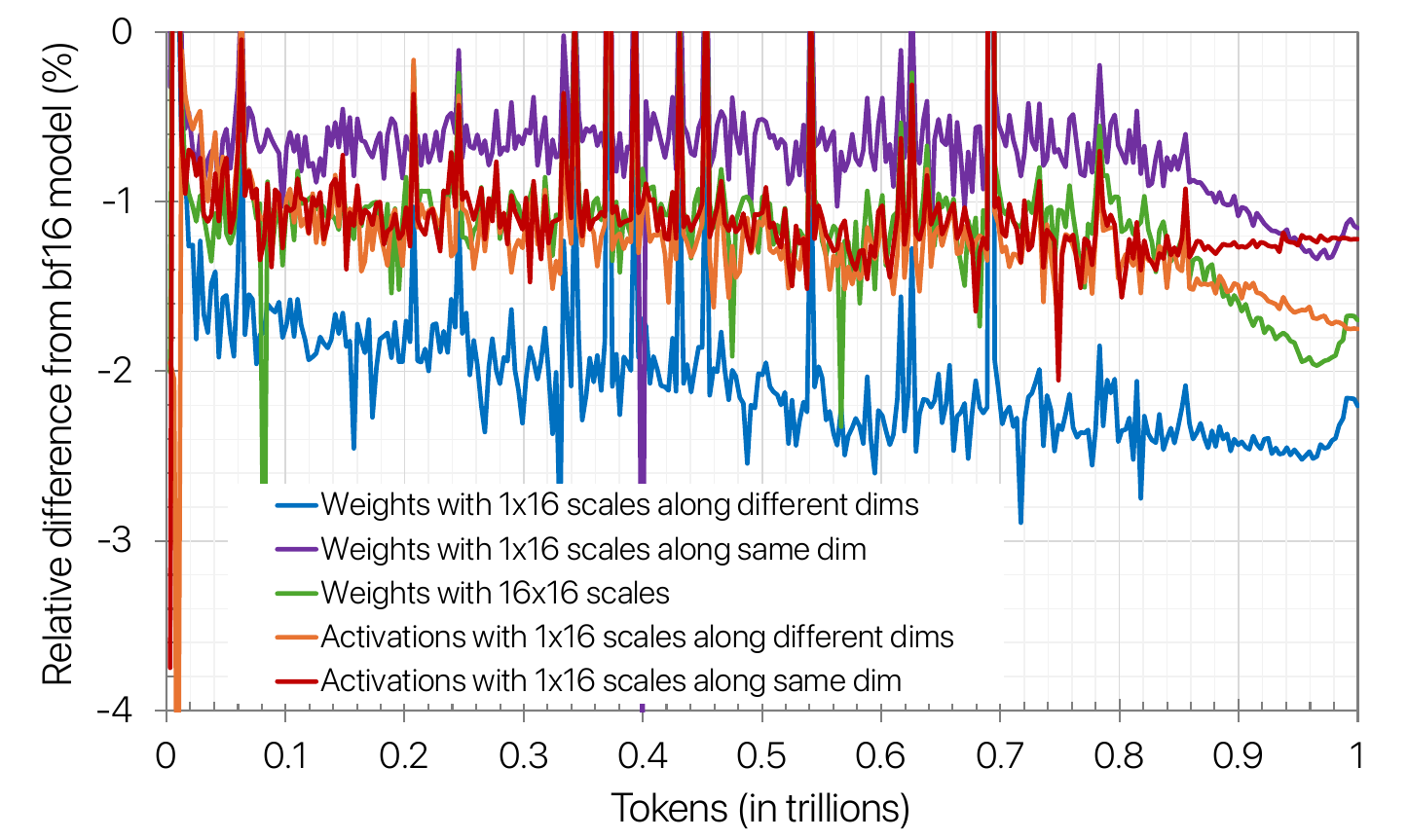}
    \caption{Effect of consistency in tensors. Relative difference in validation loss from the BF16 baseline for a 1.2B model trained on 1T tokens. NVFP4 is applied on either weights or activations. Different choices of scaling factors are applied: $1\times16$ block scales along the same dimension, $1\times16$ block scales along different dimensions, and $16\times16$ block scales, along with a global FP32 per-tensor scale.}
    \label{fig:scale_mismatch}
\end{figure}

\subsection{Consistent representations between tensors}\label{app:2d}
Applying scaling and Hadamard transforms on a weight or activation tensor typically results in different quantized representations in the forward and backward pass. We therefore study the impact of inconsistent representations for tensors during model training. In particular, we consider different choices for scale factors: (1) $1\times16$ block scales along the same dimension (i.e., input channels) in the forward and backward pass, (2) $1\times16$ block scales along different dimensions (i.e., dot-product dimension, which changes from input channels in forward to output channels in backward) , and (3) $16\times16$ block scale factors. While (1) and (3) maintain the same quantized representation in both stages of training, (2) will have different quantizations between forward and backward. Only (2) and (3) can be implemented in practice, as Tensor Cores require scaling factors along the dot-product dimension, which is transposed in the backward pass.

In Figure~\ref{fig:scale_mismatch}, we observe that having different quantized weight tensors negatively impacts the loss throughout training of the 1.2B model, where (1) achieves better accuracy than (2). Scaling using 2D blocks in (3) also improves the loss over (2), despite having a larger block granularity. On the other hand, activations are less sensitive to consistency between tensors in the forward and backward pass, and only impacted in the later stages during the learning rate decay. We hypothesize that weights are more impacted than activations because errors induced from inconsistent weights materialize in the activation gradients, which flow through the model layers during backpropagation. We also suspect that applying Hadamard transforms exacerbates the inconsistency and further impacts model accuracy.

\end{document}